\documentclass[11pt]{article}

\usepackage{times}
\usepackage{graphicx} 

\usepackage{natbib}

\usepackage{algorithm}
\usepackage{algorithmic}

\usepackage{hyperref}
\usepackage[margin=1.25in]{geometry}
\usepackage{changepage}

\title{NonSTOP: A NonSTationary Online Prediction Method for Time Series}
\author{
  Christopher Xie\\
  University of Washington\\
  \texttt{chrisxie@cs.washington.edu}
  \and
  Avleen Bijral\\
  Microsoft\\
  \texttt{avbijral@microsoft.com}
  \and
  Juan Lavista Ferres\\
  Microsoft\\
  \texttt{jlavista@microsoft.com}
}

\usepackage[english]{babel}
\usepackage[utf8]{inputenc}
\usepackage{amsmath}
\usepackage{amsthm}
\usepackage{graphicx}
\usepackage{subfig}
\usepackage{caption}
\usepackage{hyperref}
\usepackage{amssymb,listings,mathtools,gensymb,mathrsfs,bbm,bm,color,algorithm,algorithmic}
\usepackage{natbib}
\newcommand{\presec}{\vspace*{-0pt}}
\newcommand{\postsec}{\vspace*{-0pt}}
\newcommand{\presubsec}{\vspace*{-0pt}}
\newcommand{\postsubsec}{\vspace*{-0pt}}
\newcommand{\preeq}{\vspace*{-0pt}}
\newcommand{\posteq}{\vspace*{-0pt}}
\newcommand{\preitemize}{\vspace*{-0pt}}%
\newcommand{\postitemize}{\vspace*{-0pt}}%
\newcommand{\argmin}{\operatornamewithlimits{argmin}}

\newcommand{\colvec}[1]{\begin{bmatrix}#1\end{bmatrix}}

\def\mc{\mathcal}
\newcommand{\removed}[1]{}

\newcommand{\chrisnote}[1]{{  \color{red} [[ #1 -- Chris ]] }}

\newcommand{\bgamma}{\boldsymbol{\gamma}}
\newcommand{\vx}{\mathbf{x}}

\newcommand{\vy}{\mathbf{y}}
\newcommand{\vY}{\mathbf{Y}}
\newcommand{\vz}{\mathbf{z}}
\newcommand{\vZ}{\mathbf{Z}}
\newcommand{\vA}{\mathbf{A}}
\newcommand{\vF}{\mathbf{F}}
\newcommand{\vT}{\mathbf{T}}

\newcommand{\vgamma}{\boldsymbol{\gamma}}
\newcommand{\valpha}{\boldsymbol{\alpha}}
\newcommand{\vbeta}{\boldsymbol{\beta}}
\newcommand{\vpsi}{\boldsymbol{\psi}}
\newcommand{\vxi}{\boldsymbol{\xi}}

\newcommand{\vTheta}{\boldsymbol{\Theta}}
\newcommand{\vPhi}{\boldsymbol{\Phi}}

\newcommand{\sumi}[2]{\sum_{{#1}=1}^{#2}}

\newcommand{\expec}{\mathbb{E}}

\def\bal#1\eal{\begin{align*}#1\end{align*}}
\newcommand{\R}{\mathbb{R}}
\newcommand{\T}{\intercal}

\newcommand{\veps}{\varepsilon}
\newcommand{\bveps}{\boldsymbol{\varepsilon}}

\newtheorem{theorem}{Theorem}[section]
\newtheorem{remark}{Remark}[theorem]
\newtheorem{corollary}{Corollary}[theorem]
\newtheorem{lemma}{Lemma}[section]

\begin{document}
\maketitle

\begin{abstract}
We present online prediction methods for time series that let us explicitly handle nonstationary artifacts (e.g. trend and seasonality) present in most real time series.  Specifically, we show that applying appropriate transformations to such time series before prediction can lead to improved theoretical and empirical prediction performance. Moreover, since these transformations are usually unknown, we employ the learning with experts setting to develop a fully online method (NonSTOP-\textbf{NonST}ationary \textbf{O}nline \textbf{P}rediction) for predicting nonstationary time series. This framework allows for seasonality and/or other trends in univariate time series and cointegration in multivariate time series. Our algorithms and regret analysis subsume recent related work while significantly expanding the applicability of such methods. For all the methods, we provide sub-linear regret bounds using relaxed assumptions. The theoretical guarantees do not fully capture the benefits of the transformations, thus we provide a data-dependent analysis of the follow-the-leader algorithm that provides insight into the success of using such transformations. We support all of our results with experiments on simulated and real data.
\end{abstract}

\section{Introduction}
\presec

Time series modeling and forecasting is fundamentally important in many domains including econometrics and resource consumption forecasting \citep{hamilton1994time}. Analyzing and forecasting stationary time series models such as AutoRegressive Moving Average (ARMA) models \cite{BoxJenkins,BrockDavis,hamilton1994time}  has been well-studied. However, the inherently complex structure of real world data is more appropriately modeled by nonstationary time series. Time series that exhibit such nonstationary structure include seasonal time series such as influenza rates \citep{ILInet}, and time series exhibiting trends such as housing indexes and stock market prices \citep{SP}. Such data are ubiquitous and will only continue to grow as technology develops, especially with the Internet of Things where devices will generate large quantities of nonstationary time series data. Thus, efficient estimation and prediction with such models will become much more relevant.

\removed{In the analysis of time series, AutoRegressive Moving Average (ARMA) models \cite{BoxJenkins,BrockDavis,hamilton1994time} are simple and powerful descriptors of weakly stationary processes. As such, they have found tremendous application in many domains including linear dynamical systems, econometrics and forecasting resource consumption \citep{hamilton1994time}. Especially now with the internet of things (IoT),  devices will generate large quantities of time series data and thus efficient estimation and prediction with such models will become much more relevant. }

\removed{Despite a large amount of literature on estimation and prediction for these models, most of it remains within the confines of the statistical assumption of Gaussianity. Such assumptions are often unrealistic \citep{Tho94} and lead to poor prediction performance. Moreover, since the noise sequence is not known beforehand, standard methods of ARMA estimation rely on conditional likelihood estimation. These methods usually lead to nonlinear estimation problems and only hold for Gaussian residual sequences and the least squares loss.}

In the setting of streaming or high-frequency time series, one would ideally like to have methods that update the model, predict sequentially, and do not rely on any restricting assumptions on the noise sequence or the loss function. This brings attention to the paradigm of online learning \citep{cesa2006prediction}. In that vein, \citet{AnavaHazan} recently presented online gradient descent (OGD) and online Newton step (ONS) methods (ARMA-OGD and ARMA-ONS) for ARMA prediction that do not make the Gaussianity assumption. Using a truncated auto-regressive (AR) representation of an ARMA process, the authors provide online ARMA prediction algorithms with sublinear regret, where the regret is with respect to the best conditionally expected one-step ARMA prediction loss in hindsight. While no assumption is made about the stationarity of the generating ARMA process, the empirical performance of ARMA-OGD suffers in the presence of seasonality and/or trends \citet{liu2016online}. 

To handle a deterministic or stochastic trend, \citet{liu2016online} recently presented ARIMA-OGD, a straigtforward extension of ARMA-OGD using AutoRegressive Integrated Moving Average (ARIMA) models. However, the trend transform and its parameters (e.g. order of integration) are assumed to be known, which is unrealistic in online settings as one typically needs adequate data to test for such nonstationarities. \removed{Also, a fixed transform may not adapt well to the changes in the incoming data.}Moreover, these methods don't account for seasonality and  do not carry over to the multivariate domain. These shortcomings of existing work necessitate the development of broader methods that take into account different types of nonstationarities with extensions to multivariate time series.

\postsubsec
\subsection{Contributions}
\presubsec

We provide general methods for time series prediction using OGD \citep{Zinkevich2003OnlineCP} that account for possible nonstationarities in the data. This leads to explicit transformations of the data before prediction when the form of these nonstationarities are known. In the univariate case, our approach subsumes existing work while expanding the applicability of such online methods to more realistic time series settings. For the multivariate case, we propose a novel algorithm for prediction of nonstationary vector time series generated by Error Corrected Vector AutoRegressive Moving Average (EC-VARMA) processes to deal with the phenomenon of cointegration \citep{tsay2013multiple,lutkepohl2005new}. Estimating EC-VARMA models are non-trivial in general; this typically requires a two-stage process where the cointegrating rank is estimated before the parameters are estimated. The algorithm we propose simultaneously estimates both the cointegrating rank and the VARMA (Vector AutoRegressive Moving Average) matrix parameters. 

However, the form of the nonstationary transformations are usually unknown. These transforms are typically determined by statistical tests on a fixed dataset with sample size requirements. In the online setting, this is unrealistic, thus we unify the above methods into a meta-algorithm called NonSTOP to learn the correct transformation in an online fashion. NonSTOP utilizes the weighted majority method \cite{cesa2006prediction} wherein each expert corresponds to different parameter settings of the nonstationary transformation (e.g. trend only, trend and seasonality, no trend/seasonality, etc). NonSTOP quickly hones in on the correct transformation, and also allows for flexibility in adapting to changes in the data. 

Our regret analysis, which provides sublinear regret guarantees for all methods, only requires invertibility of the moving average polynomial while the assumptions in \cite{AnavaHazan} and \cite{liu2016online} are less natural. Moreover, we don't require an upper bound on the data as nonstationary data can be unbounded.

To emphasize the effect of the these nonstationary transformations, we prove a data dependent regret guarantee for FTL (for least squares loss) that gives insights into why adjusting for nonstationarities can give faster convergence.

All proofs can be found in the supplement.

\subsection{Related Work}

The application of online learning to time series modeling has begun to receive more attention in the past couple of years. \citet{AnavaHazan} developed online gradient and online second order methods for ARMA prediction. \citet{liu2016online} present a trend extension to \cite{AnavaHazan} that requires knowledge of the parameters of the trend transformation, which is unrealistic in the online setting. Other extensions include application to missing data \cite{anava2015online} and heteroscedastic processes \cite{anava2016heteroscedastic}. However, these algorithms suffer in the presence of seasonliaty and are not applicable to multivariate time series. In contrast, we provide a unified general framework capable of efficiently handling many common types of nonstationarities found in real data.

Few works have tackled the problem of forecasting nonstationary time series in an online fashion. \citet{Vitaly2016} developed generalization bounds for nonstationary time series prediction by using online-to-batch conversion techniques on the sequences of hypothesis output by a online algorithm for time series prediction. However, the work develops guarantees for the batch setting as opposed to the online (possibly adversarial) setting and the method presented is in general computationally intensive. Recently, \citet{hazan2017learning} presented an online prediction method for a linear dynamical system and also provided optimal regret bounds. Even though general state space models allow for nonstationary components, these were not explored in the work. While ARMA models do have a linear dynamical system representation, their natural form is more parsimonious for explicitly modeling different kinds of nonstationarity.

\postsec

\section{Preliminaries: Time Series Modeling}
\label{sec:preliminaries}
\presec

In this section, we provide a brief summary of SARIMA (Seasonal ARIMA) and EC-VARMA processes. For more comprehensive background, see \cite{BoxJenkins, tsay2013multiple}.
\removed{seasonal and/or integrated extensions to standard ARMA processes. We also provide a brief introduction to VARMA and EC-VARMA processes. }For an introduction to online convex optimzation and online gradient descent, please see \citep{shalev2011online, Zinkevich2003OnlineCP, hazan2007logarithmic}. We introduce standard notation for time series in Table \ref{table:notation}. Note that the differencing notation $\Delta$ can be compounded: $\Delta^2 x_t = \Delta (x_t - x_{t-1}) = x_t - 2x_{t-1} + x_{t-2}$.

\begin{table}[t]
  \caption{Notation}
  \label{table:notation}
  \centering
\begin{tabular}{|c|c|}
    \hline
    $B$ & $Bx_t = x_{t-1}$ \\ \hline
    $\Delta$ & $\Delta x_t = x_t - x_{t-1}$  \\ \hline
    $\Delta_s$ & $\Delta_s x_t = x_t - x_{t-s}$ \\ \hline
    $d$ & Differencing order \\ \hline
    $s$ & Seasonal period \\ \hline
    $\tilde{D}$ & Seasonal differencing order \\ \hline
    
\end{tabular}
\end{table}

\postsubsec
\subsection{SARIMA}
\label{sec:sarima_processes}
\presubsec

Time series exhibiting seasonal patterns can be modeled by Seasonal AutoRegressive Integrated Moving Average (SARIMA) Processes. Let $x_t, \veps_t \in \R$ denote the time series and innovations (random variables). SARIMA$(p,d,q) \times (P,\tilde{D},Q)_s$ processes are described by the following:
\preeq
\begin{equation} \label{eq:sarima_lag_poly_form}
\phi(B) \Phi(B^s) \Delta^d \Delta^{\tilde{D}}_s x_t = \theta(B) \Theta(B^s) \veps_t
\end{equation}
\posteq
where $\phi(B) = 1 - \sumi{i}{p} \phi_i B^{i}, \theta(B) = 1 + \sumi{i}{q} \theta_i B^{i}$, $\Phi(B^s) = 1 - \sumi{i}{P} \Phi_i B^{is}, \Theta(B^s) = 1 + \sumi{i}{Q} \Theta_i B^{is}$ and $\phi, \Phi, \theta, \Theta \in \R$. \removed{$\phi(B)$ and $\theta(B)$ are the non-seasonal autoregressive (AR) and moving average (MA) lag polynomials, respectively. Similarly, $\Phi(B^s)$ and $\Theta(B^s)$ are the seasonal AR and the seasonal MA lag polynomials, respectively.}
Note that $\tilde{D}=0$ implies a ARIMA($p,d,q$) process, and $\tilde{D}=d=0$ implies a ARMA($p,q$) process.

SARIMA processes explicitly model trend and seasonal nonstationarities by assuming that the differenced process $\Delta^d \Delta^{\tilde{D}}_s x_t$ is an ARMA process with AR lag polynomial $\phi(B)\Phi(B^s)$ and MA lag polynomial $\theta(B)\Theta(B^s)$. We denote the order of the underlying AR and MA lag polynomials as $l_a$ and $l_m$, respectively. For SARIMA$(p,d,q)\times (P,\tilde{D},Q)_s$ processes, Eq. (\ref{eq:sarima_lag_poly_form}) gives us that $l_a = p+Ps$ and $l_m = q+Qs$.

If the MA lag polynomial has all of its roots outside of the complex unit circle, then the SARIMA process is defined as invertible. \removed{Let $\beta_i$ be the scalar coefficients of the MA lag polynomial. - MOVE TO SUPPLEMENT} Invertibility is equivalent to saying that the companion matrix (see Supplement)
\removed{\begin{equation} \label{eq:companion_matrix}
\vF = \colvec{-\beta_1 & -\beta_2 & \ldots & \ldots & -\beta_{l_m} \\ 1 & 0 & \ldots & \ldots & 0 \\ 0 & 1 & 0 & \ldots & \vdots \\ \vdots & 0 & \ddots & \ddots & \vdots \\ 0 & \vdots & \vdots & 1 & 0}
\end{equation} - MOVE TO SUPPLEMENT}
has eigenvalues less than 1 in magnitude. If this is the case, then the underlying ARMA process $\Delta^d \Delta^{\tilde{D}}_s x_t$ can be written as an AR$(\infty)$ process and can be approximated by a finite truncated AR process. 

\removed{\subsection{VARMA Processes}
Vector AutoRegressive Moving Average (VARMA) processes provide a parsimonious description of modeling linear multivariate time series.
\removed{The generalization of ARMA processes to the multivariate domain is called vector ARMA (VARMA) models.} Let $\vx_t, \bveps_t \in \R^k, \vPhi_i \in \R^{k \times k}, \vTheta_i \in \R^{k \times k}$. A VARMA$(p,q)$ process is described by:
\begin{equation} \label{eq:varma}
\vx_t = \sumi{i}{p} \vPhi_i \vx_{t-i} + \sumi{i}{q} \vTheta_i \bveps_{t-i} + \bveps_t
\end{equation}
which can also be written in lag polynomial form:
\begin{equation} \label{eq:varma_lag_poly_form}
\vPhi(B) \vx_t = \vTheta(B) \bveps_t
\end{equation}
with $\vPhi(B) = I - \sumi{i}{p} \vPhi_i B^i, \vTheta(B) = I + \sumi{i}{q} \vTheta_i B^i.$ The requirements for invertibility are very similar to the univariate case. We require that $\det\left(\vTheta(B) \right)$ must have all of its roots outside of the complex unit circle. Again, this is equivalent to saying that the companion matrix has eigenvalues less than 1 in magnitude \citep{lutkepohl2005new, tsay2013multiple}. If the process is invertible, then it can be rewritten as a VAR$(\infty)$ process. - MOVE TO SUPPLEMENT}

\postsubsec
\subsection{EC-VARMA}
\presubsec

In many cases, a collection of time series may follow a common trend. This phenomenon, known as cointegration, is ubiquitous in economic times series \citep{tsay2013multiple}. Let $\vx_t, \bveps_t \in \R^k, \vPhi_i \in \R^{k \times k}, \vTheta_i \in \R^{k \times k}$. First, a  VARMA$(p,q)$ process is described by:
\preeq
\begin{equation} \label{eq:varma}
\vx_t = \sumi{i}{p} \vPhi_i \vx_{t-i} + \sumi{i}{q} \vTheta_i \bveps_{t-i} + \bveps_t
\end{equation}
\posteq
which is equivalent to writing $\vPhi(B)\vx_t = \vTheta(B)\bveps_t$ where $\vPhi(B) = I - \sumi{i}{p} \vPhi_i B^i, \vTheta(B) = I + \sumi{i}{q}\vTheta_i B^i$. Formally, $\vx_t$ is cointegrated if $\Delta \vx_t$ is stationary and there exists a vector $\mu \in \R^k$ such that $\mu^\T \vx_t$ is a stationary process. If $\vx_t$ is cointegrated, then we can rewrite the original VARMA representation of $\vx_t$ as
\preeq
\begin{equation} \label{eq:ec_varma_form}
\Delta \vx_t = \Pi \vx_{t-1} + \sumi{i}{p-1} \Gamma_i \Delta \vx_{t-i} +  \sumi{i}{q} \vTheta_i \bveps_{t-i} + \bveps_t
\end{equation}
\posteq
where $\Pi = \vPhi(1)$, denoted the cointegrating matrix, is low rank (cointegrating rank), and $\Gamma_j = -(\vPhi_{j+1} + \ldots + \vPhi_p)$ for $ j = 1,\ldots p-1$. Eq. (\ref{eq:ec_varma_form}) is known as an Error-Corrected VARMA (EC-VARMA) model. Given that an such a process starts at some fixed time $t=0$ with fixed initial values, we can write Eq. (\ref{eq:ec_varma_form}) in a pure EC-VAR form \cite{lutkepohl2006forecasting}:
\begin{equation}
\Delta \vx_t = \Pi^* \vx_{t-1} + \sumi{i}{t-1} \Gamma_i^* \Delta \vx_{t-i} + \bveps_t,\ \ \ t \in \mathbb{N}
\end{equation}
This allows us to approximate an EC-VARMA process with an EC-VAR model.

\postsec

\section{Univariate Methods}
\label{sec:online_time_series_prediction_framework}
\presec

In this section, we present algorithms for online univariate time series prediction which subsumes recent works such as ARMA-OGD as presented in \cite{AnavaHazan} and its the extension to trend nonstationarities ARIMA-OGD as presented in \cite{liu2016online}.  

We show that time series with certain characteristics (such as a trend and/or seasonality) can be transformed before prediction to give better theoretical and empirical results. To achieve this goal, we present a unified template for {\textbf T}ime {\textbf S}eries{\textbf{ P}rediction using OGD, denoted TSP-OGD, that allows for prediction of transformed time series.  The choice of the transformation, dependent on the underlying data generation process (DGP), can lead to improved regret guarantees, partially explaining why these transformations lead to better empirical performance.

This framework includes some commonly used transformations \removed{of seasonal and non-seasonal differencing} \citep{BoxJenkins}. Table \ref{table:dgp} shows the explicit form of such transformations. In practice, the order of differencing is usually determined by statistical tests (e.g. \cite{EJT1996}) on a given dataset, which is not realistic when considering the online setting.

\postsubsec
\subsection{TSP-OGD}
\presubsec

We assume the following for the remainder of this section:
\preitemize
\begin{itemize}
\item[U1)] $x_t$ is generated by a DGP such that there exists a transformation $\tau(x_t)$ which results in an invertible ARMA process. Moreover, there corresponds an inverse transformation $\zeta$ that satisfies $\zeta(\tau(x_t)) = x_t$. Examples of such processes are ARMA, ARIMA, and SARIMA processes.

\item[U2)] The noise sequence $\veps_t$ of the process is independent. Also, it satisfies that $\expec [|\veps_t|] < M_{\textrm{max}} < \infty$. 

\item[U3)] $\ell_t: \R^2 \rightarrow \R$ is a convex loss function with Lipschitz constant $L > 0$.

\item[U4)] We assume the companion matrix $\vF$ (as defined in the Supplement) of the MA lag polynomial is diagonalizable, i.e. $\vF = \vT \Lambda \vT^{-1}$ where $\Lambda$ is a diagonal matrix of eigenvalues. Denote $\lambda_{\max}$ as the magnitude of the largest eigenvalue ($\lambda_{\max} < 1$ by definition of invertibility), and $\kappa \in \R$ such that $\left (\sigma_{\max}(\vT)/\sigma_{\min}(\vT)\right) \leq \kappa$.

\end{itemize}
\postitemize
Assumption U1 includes a large class of models including ARMA/ARIMA/SARIMA models. It is well-known that the class of ARMA models is equivalent to linear state space models, thus that class of models is included in U1. U2 and U3 are standard assumptions in the time series literature. Lastly, U4 is a relaxation of the less natural assumptions present in previous works \cite{AnavaHazan, liu2016online}. Our assumption only requires the invertibility of the MA process, which guarantees that the process can be well-approximated by a finite AR process (which is the heart of our framework). Note that we don't make the assumption that the data is bounded (as in previous works), as data generated by a nonstationary process can be unbounded (e.g. random walks).

\begin{algorithm}[t]
\caption{TSP-OGD Framework}
\label{alg:tsp_ogd}
\begin{algorithmic}[1]
\REQUIRE DGP parameters $l_a, l_m$. Horizon $T$. Learning rate $\eta$. Data: $\{x_t\}$. Transformation $\tau$. Inverse Transformation $\zeta$.
\STATE Set $M = \log_{\lambda_{\max}} \left(\left( 2\kappa TLM_{\max}\sqrt{l_m} \right)^{-1} \right) + l_a $
\STATE Transform $x_t$ to get $\tau(x_t)$.
\STATE Choose $\bgamma^{(1)} \in \mathcal{E}$ arbitrarily.
\FOR {$t=1$ to $T$}
\STATE $\tau\left(\tilde{x}_t\right) = \sumi{i}{M} \bgamma^{(t)}_i \tau\left( x_{t-i}\right)$
\STATE Predict $\tilde{x}_{t} = \zeta(\tau(\tilde{x}_t))$
\STATE Observe $x_t$ and receive loss $\ell_t^M\left(\bgamma^{(t)}\right)$
\STATE Set $\bgamma^{(t+1)} = \Pi_\mathcal{E}\left( \bgamma^{(t)} - \eta \nabla \ell_t^M\left( \bgamma^{(t)} \right) \right)$
\ENDFOR
\end{algorithmic}
\end{algorithm}

\begin{table}[t]
  \caption{DGPs and their Transformations}
  \label{table:dgp}
  \centering
\begin{tabular}{|c|c|c|}
    \hline
    DGP & $\tau(x_t)$ &  $\zeta(y_t)$ \\ \hline
    ARMA & $x_t$ &  $y_t$ \\ \hline
    ARIMA & $\Delta^d x_t$ & $ y_t + \sum_{i=0}^{d-1} \Delta^i x_{t-1}$ \\ \hline
    SARIMA & $\Delta^d\Delta^{\tilde{D}}_s x_t$ & $ \begin{array}{lcl} y_t &+ &\sum_{i=0}^{d-1} \Delta^i \Delta_s^{\tilde{D}} x_{t-1}\\ &+ &\sum_{i=0}^{{\tilde{D}}-1} \Delta_s^i x_{t-s} \end{array}$ \\ \hline
\end{tabular}
\end{table}

In Algorithm \ref{alg:tsp_ogd}, the model parameters of the stochastic process are fixed by an adversary. At time $t$, $\veps_t$ and $x_t$ are generated by the DGP. Before $x_t$ is revealed to us, the learner makes a prediction $\tilde{x}_t$ (see line 6 of Algorithm \ref{alg:tsp_ogd})  which incurs a prediction loss of $\ell_t(x_t, \tilde{x}_t)$. In more detail, this prediction is preceded by a transform $\tau$ (See Table \ref{table:dgp}) that may require data points from previous rounds (we suppress that dependence in the notation for convenience).  The prediction $\tau(\tilde{x}_t) := \sumi{i}{M} \gamma_i \tau\left( x_{t-i} \right)$ is computed using an AR model of order $M$ to approximate the underlying invertible ARMA process. Then it is inverted with $\zeta$ and incurs a loss
\preeq
\begin{align} \label{eq:approx_loss}
\ell_t^M(\bgamma) &:= \removed{\ell_t\left( x_t, \zeta\left( \tau(\tilde{x}_t) \right) \right) \notag \\
&=} \ell_t\left(x_t, \zeta\left( \sumi{i}{M} \gamma_i \tau\left( x_{t-i} \right) \right) \right)
\end{align}
\posteq
where $\bgamma$ is the vector of parameters of the approximating AR model. \removed{The regret needs to be written w.r.t the most general Multiplicative Seasonal Arima model, since you can get arma,arima,sarima regret by setting d,s,D etc.}The prediction performance is evaluated using an ``extended'' notion of regret that looks at the prediction loss of the best process in hindsight. \removed{The prediction loss of a DGP uses the conditional expectation of the underlying ARMA process, then applies the inverse transform $\zeta$.} Precisely, let $\valpha, \vbeta$ denote the set of AR and MA parameters, respectively, of the underlying ARMA process $\tau(x_t)$. Define
\preeq
\begin{equation} \label{eq:transformed_arma_loss}
f_t(\valpha, \vbeta) = \ell_t \left(x_t, \zeta\left( \expec\left[\tau(x_t) |  \{ \tau\left( x_t \right) \}_{t=1}^{t-1}; \valpha, \vbeta \right] \right) \right)
\end{equation}
\posteq
Note that $f_t$ depends on the transformations $\tau,\zeta$ in U1. The extended regret is defined as comparing the accumulated loss in Eq. (\ref{eq:approx_loss}) to the loss of the best process in hindsight:
\preeq
\begin{equation} \label{eq:univariate_extended_regret}
\text{Regret} = \sumi{t}{T} \ell_t^M (\vgamma^{(t)}) - \min_{\valpha, \vbeta \in \mathcal{K}} \sumi{t}{T} \expec [f_t(\valpha, \vbeta)]
\end{equation}
\posteq
where $\mathcal{K}$ is the set of invertible ARMA processes. \removed{Note that the randomness in the expectation is w.r.t. the noise sequence $\veps_t$ while the data $x_t$ is fixed.}

Furthermore, let $\mathcal{E} \subseteq \R^M$ be a convex set of approximating AR models, i.e. $\vgamma \in \mathcal{E}$. $\mathcal{E}$ should be chosen to be large enough to include a valid approximation to the DGP described in U1. However, since the DGP is unknown in practice, one usually chooses a simple constraint set such as $\mathcal{E} = \{\vgamma : \|\vgamma\|_\infty \leq 1 \}$. Let $D = \sup_{\vgamma_1, \vgamma_2 \in \mathcal{E} } \|\vgamma_1 - \vgamma_2\|_2$, and $\|\nabla_{\vgamma} \ell_t^M(\vgamma)\|_2 \leq G(t)$ for some monotonically increasing $G(t)$. 
This assumption allows the time series to be potentially unbounded. \removed{As an example, the norm of the gradient for the squared loss depends on the bound on the data.} Let $\Pi_\mathcal{E}$ denote the projection operator onto the set $\mathcal{E}$. 

 We present a general regret bound for Algorithm \ref{alg:tsp_ogd}:

\begin{theorem} \label{thm:tsp_ogd} Let $\eta = \frac{D}{G(T)\sqrt{T}}$. Then for any data sequence $\{x_t\}_{t=1}^T$ that satisfies assumptions U1-U4, Algorithm \ref{alg:tsp_ogd} generates a sequence $\{\vgamma^{(t)}\}$ in which
\preeq
\begin{equation*}
\text{Regret} \removed{\sumi{t}{T} \ell_t^M (\vgamma^{(t)}) - \min_{\valpha, \vbeta \in \mathcal{K}} \sumi{t}{T} \expec [f_t(\valpha, \vbeta)]} = O\left( DG(T) \sqrt{T} \right)
\end{equation*}
\posteq
\end{theorem}
\emph{Remark 1:} Note that plugging in the ARMA transformation and ARIMA transformation in Table \ref{table:dgp} to Algorithm \ref{alg:tsp_ogd} recovers ARMA-OGD as presented in \cite{AnavaHazan} and ARIMA-OGD as presented in \cite{liu2016online}, respectively. Plugging in the SARIMA transformation results in a novel variation which we denote as SARIMA-OGD.

For the following remarks, assume that $\ell_t$ is squared loss, the DGP is a SARIMA process, and $|x_t| < C(t)=O\left(\log t\right)$ (note that the $\log$ transformation is commonly employed as a variance stabilizer in many time series domains).

\emph{Remark 2:} Table \ref{table:univariate_regret_comparison} shows the regret bounds obtained by using different transformations/algorithms. The differencing transforms remove any growth trends in the data; as a consequence the transformed time series is bounded by a constant. In our case, this implies $|\Delta^d x_t|, |\Delta^d \Delta^{\tilde{D}}_s x_t| < C_\Delta$ (a constant), which leads to an improvement over the regret bound obtained from ARMA-OGD. This improvement can be seen in the empirical results section of \cite{liu2016online}.

\emph{Remark 3:} When the DGP is assumed to be SARIMA, we require that $l_a = p+Ps, l_m = q+Qs$ as mentioned in Section \ref{sec:preliminaries}, i.e. $l_a,l_m$ both need to essentially be a multiplicative factor larger than $s$. This affects the length of the required AR approximation $M$ as described in line 1 of Algorithm \ref{alg:tsp_ogd}.

\removed{\emph{Remark 4:} Table \ref{table:univariate_regret_comparison} suggests that using the ARIMA transformation gives the same regret as when using the SARIMA transformation. However, in this case the SARIMA transformation empirically (Section \ref{sec:empirical_results}) outperforms the ARIMA transformation.  As such, we believe that these results do not paint a complete picture as to why we achieve faster empirical convergence and a more data dependent phenomenon is perhaps at play here. In Section \ref{sec:data_dependent_regret_bounds}  we explore a data dependent analysis of FTL.}

\removed{\begin{corollary}
Assume $|\Delta^d \Delta^d_s x_t| < C_\Delta$. Then plugging in a SARIMA process as the DGP to Theorem \ref{thm:tsp_ogd} results in this regret bound:
\[
O\left(DG(T)\sqrt{T}\right) = O\left( C_\Delta^2 M^2 \sqrt{T} \right)
\]
\end{corollary}

\begin{remark}
Our theorem encompasses Theorem 3.5 as presented in \cite{AnavaHazan}; simply plug in ARMA as the DGP in U1. Likewise, Theorem 2 in \cite{liu2016online} can be recovered by plugging in an ARIMA process as the DGP.
\end{remark}
Note that for the regret to be sublinear, we require that $G(T) = o\left(\sqrt{T}\right)$. The proof of the theorem is in the appendix.

\postsubsec
\subsection{Accounting for Non-Stationarities}
\label{subsec:accounting_for_non_stationarities}
\presubsec

We now show that accounting for the appropriate non-stationarities in the form of transformation functions $\tau$ will result in theoretically faster convergence guarantees. We make the following additional assumptions that hold for the rest of this section:

\begin{itemize}
\item The DGP in assumption U1 is a SARIMA process with $D,d,s$ known.

\item $|x_t| < C(t) = O(\log(t))$ and $|\Delta^d x_t|, |\Delta^d \Delta^D_s| < C_\Delta$. These assumptions are justifiable since log transformation of time series is pretty standard. Also, seasonal and/or non-seasonal differencing usually takes care of explosive growth in the data.

\end{itemize}

}

\begin{table}[t]
  \caption{Regret Bounds for Different Transformations}
  \label{table:univariate_regret_comparison}
  \centering
\begin{tabular}{|c|c|c|}
    \hline
     Algorithm & $\tau(x_t)$ & Regret Bound  \\ \hline
     ARMA-OGD & $x_t$ & $O\left(M^2 \log^2 (T)  \sqrt{T} \right)$  \\ \hline
     ARIMA-OGD & $\Delta^d x_t$ & $O\left( M^2 \sqrt{T} \right)$  \\ \hline
     SARIMA-OGD & $\Delta^d \Delta^{\tilde{D}}_s x_t$ & $O\left(M^2  \sqrt{T} \right)$ \\ \hline
\end{tabular}
\end{table}

\removed{
With assumption U5, the MA lag polynomial is described as $\theta(L)\Theta(L^s)$ as in Eq. \ref{eq:sarima_lag_poly_form}. In order for Theorem \ref{thm:tsp_ogd} to hold, we need to set $l = q+Qs$, i.e. $l$ needs to essentially be a multiplicative factor larger than $s$. This affects the size of $m$ as described in line 1 of Algorithm \ref{alg:tsp_ogd}, which is of the practitioners interest. Note that the coefficients of $\theta(L)\Theta(L^s)$ correspond to $\vbeta$.

\remove{Recall from Section \ref{sec:sarima_processes} that a SARIMA process can also be viewed as an ARIMA process, which can in turn be viewed as an ARMA process. This means that although the data is assumed to be generated from a SARIMA process, it is equivalent to say it is generated from an ARIMA process, or an ARMA process. Thus, we can plug in ARMA, ARIMA, or SARIMA as the DGP into Algorithm \ref{alg:tsp_ogd} and Theorem \ref{thm:tsp_ogd} will hold. }

Letting $\ell_t(x,y) = \frac12 (x-y)^2$, we plug in these three different DGPs to Algorithm \ref{alg:tsp_ogd}, we essentially plug in different transformations and compare how the algorithm fares when accounting for non-stationarities. The resulting regret bounds are shown in Table \ref{table:univariate_regret_comparison}. These results show that accounting for non-stationarities, which essentially gets rid of the correlations in the data, results in faster convergence. We see that not accounting for any non-stationarity incurs an extra $\log^2(T)$ factor in the regret, which is undesirable. This stems from the fact that $G(T) = O(\log^2(T))$ due to the assumption that $|x_t| = O(\log(t))$; when dealing with the differenced data, $G(T)$ is just a constant. However, we expect that the SARIMA transformation should outperform the ARIMA transformation since the true DGP is a SARIMA process, but this is not reflected in the bounds in Table \ref{table:univariate_regret_comparison}. As such, we believe that these results do not paint a complete picture as to why we are achieving faster empirical convergence as shown in Section \ref{sec:empirical_results}. Section \ref{sec:data_dependent_regret_bounds} delves into a possible alternative explanation.}

\removed{Next, we present a nontrivial extension of the framework to the case of multivariate nonstationary time series.} \removed{ and present a $\mc{O}(\sqrt{T})$ regret algorithm. }

\subsection{Data Transformation Dependent Regret }
\label{sec:data_dependent_regret_bounds}

The transformations discussed in the previous sections essentially diminish serial correlation in the data due to any existing nonstationarities. However, our regret bounds (shown in Table \ref{table:univariate_regret_comparison}) do not accurately reflect this. We conjecture that these bounds are missing data-dependent terms that capture correlations \removed{inherent in many} in nonstationary time series. \removed{\textcolor{red}{The stronger the correlation or dependence on the data, the slower the learning may be. For example, if $x_{t+1}$ is a slight variation of $x_t$, then we aren't getting too much more information. More variety in the data would possibly speed up the learning process. To support this intuition, we provide a data dependent guarantee for the FTL algorithm for the case of least squares.}}To give a flavor of what a satisfactory data dependent regret bound might look like, we analyze the regret for the FTL algorithm for the case of least squares loss  $\left (\ell_t(\vgamma) = \frac12 (x_t - \vgamma^\T \vpsi_t)^2 \right )$
\removed{
\begin{align} \label{eq:ls_loss}
\ell_t(\vgamma) = \frac12 (x_t - \vgamma^\T \vpsi_t)^2
\end{align}
}
 and show that these bounds depend on a data dependent term.  We look at the standard notion of regret and hence the result in this section is much more general than time series prediction and is also relevant to general regression problems.

The FTL algorithm follows a simple update \citep{shalev2011online} $\left( \vgamma_{t+1} \in \argmin_{\vgamma} \sumi{i}{t} \ell_t(\vgamma) \right)$.
\removed{
\begin{equation} \label{eq:ftl_update}
\vgamma_{t+1} \in \argmin_{\vgamma} \sumi{i}{t} \ell_t(\vgamma)
\end{equation}
}
\removed{Plugging Eq. \ref{eq:ls_loss} in Eq. \ref{eq:ftl_update}} It is easy to see that the FTL algorithm for least squares loss is just recursive least squares (RLS). Using the relevant RLS update equations \cite{ljung1998system,lai1982least}, we obtain the following:
\begin{theorem} \label{thm:ftl_data_dependent}
Let $\ell_t(\vgamma)$ be the squared loss with Lipschitz constant $L > 0$. Then FTL generates a sequence $\{\vgamma_t\}$ in which
\preeq
\[
\sumi{t}{T} \ell_t\left(\vgamma_t\right) - \min_{\vgamma} \sumi{t}{T} \ell_t \left(\vgamma \right) = O\left(\sumi{t}{T} \frac{1}{t \lambda_{\min}(t)} \right)
\]
\posteq
where $\lambda_{\min}(t) := \lambda_{\min}\left( \frac{1}{t} \sumi{i}{t} \vpsi_i \vpsi_i^\intercal \right)$.
\end{theorem}

\removed{For the transformations in Table \ref{table:dgp}, note that $x_t - \tilde{x}_t = \zeta(\tau(x_t)) - \zeta(\tau(\tilde{x}_t)) = \tau(x_t) - \tau(\tilde{x}_t)$. Thus, we can apply FTL and Theorem \ref{thm:ftl_data_dependent} to approximate ARIMA and SARIMA processes by simply using $\ell_t(\vgamma) = \frac12 \left(\tau(x_t) - \vgamma^\T \tau(\vpsi_t)\right)^2$.}

At the heart of our framework in Algorithm \ref{alg:tsp_ogd}, we are approximating an ARMA process with an AR model. In order to apply Theorem \ref{thm:ftl_data_dependent} to our time series prediction setting for DGPs as described in assumption U1, assume for now that we use FTL and least squares loss to predict the underlying ARMA process $\tau(x_t)$ with an AR model $\vgamma^\T \tau(\vxi_t)$, where $\vxi_t = \colvec{x_{t-1} & \ldots & x_{t-M}}^\T$ and $\tau(\vxi_t) =  \colvec{\tau(x_{t-1}) & \ldots & \tau(x_{t-M})} ^\T$. This results in $\lambda_{\min}(t) = \left(\frac1t \sumi{i}{t} \tau(\vxi_i) \tau(\vxi_i)^\T \right)$\removed{, which is the empirical non-centered autocovariance of the transformed data}. Ideally, we want this quantity to be large, \removed{which implies that each direction has a lot of information in it. If this quantity is small, then there are directions where the variance of $\tau(\vxi_t)$ is small,} meaning that the invidividual samples $\tau(x_t)$ are not highly correlated.

\removed{We note that \cite{Vitaly2016} recently presented data dependent generalization bounds for nonstationary time series that yield a data discrepancy measure. While their results are very general, it is not clear how this term relates to the various transformations we presented and requires solving an SDP. In contrast the smallest eignevalue of the auto-covariance is easily computed and clearly demonstrates the benefits of choosing and applying these transformations.}

To empirically assess the regret bound when accounting for the appropriate nonstationarities, we calculate the bound $\sumi{i}{T} 1/\left(t \lambda_{\min}(t)\right)$ for the three transforms in Table \ref{table:dgp}. We simulated a SARIMA process $50$ times with $T=10,000$ and then averaged the regret bound across the 50 simulated datasets using each transformation. The result is shown in Figure \ref{fig:data_dependent_regret_bounds}. The transformations essentially decrease correlations making the data more like realizations of a stationary process; we can see that accounting for the appropriate nonstationarities results in tighter regret bounds.

\begin{figure}[t]
  \centering
\scalebox{1.0}{
  \includegraphics[width=2.5in]{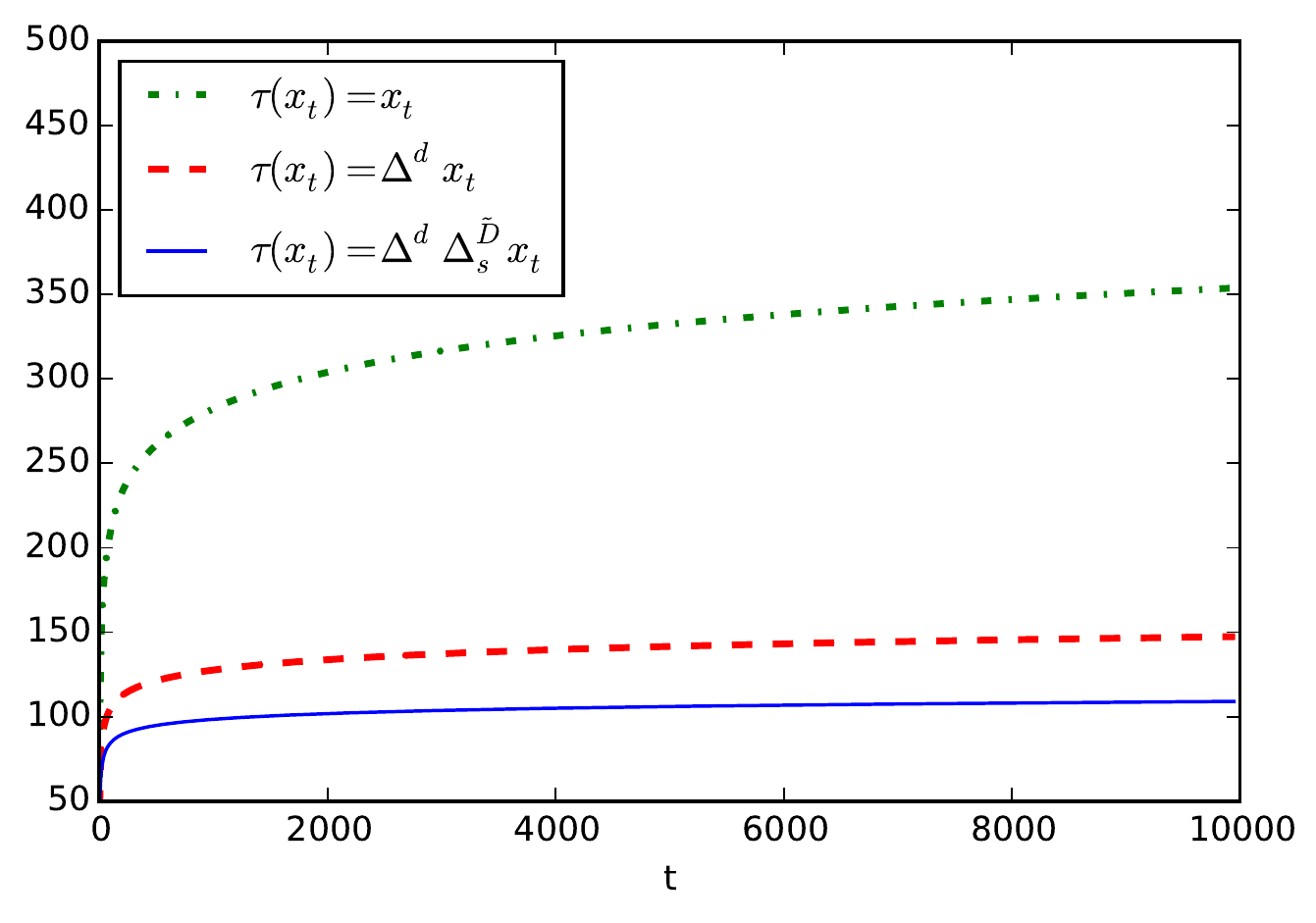}}
  \caption{Empirical Regret Bounds}
  \label{fig:data_dependent_regret_bounds}  
\end{figure}

\removed{\chrisnote{Maybe a sentence mentioning future work involves searching for data-dependent bounds for OGD algorithms.}}

\removed{Missing from the analysis present in this section and other related works is the case of multivariate time series. The issue of nonstationarity in such series is complicated by the fact that differencing transforms do not preserve the conintegrating structure prevalent in many real time series. Thus, online methods must additionally infer this relationship from the data.}

\postsec

\section{Multivariate Methods}
\label{sec:error_corrected_varma_models}
\presec

Online prediction using multivariate nonstationary models present an additional difficulty due to the notion of cointegration (Section \ref{sec:preliminaries}). Estimating EC-VARMA models in the static setting is non-trivial in general since the cointegrating rank is unknown and is typically determined by statistical tests (e.g. trace statistic of \citet{Johansen88}), which again is not realistic in the online setting. We propose a novel online method for cointegrated vector time series that simultaneously updates both the cointegrating matrix (including its rank) and the approximating VAR matrix parameters in order to accurately adapt to the underlying DGP and make predictions. 

\removed{\textcolor{red}{ALTERNATIVE: Note that by straight-forwardly generalizing assumptions U1-U6 and Algorithm \ref{alg:tsp_ogd} to the multivariate domain, we can obtain a multivariate extension of Theorem \ref{thm:tsp_ogd}. We are able to plug in VARMA as the assumed DGP to obtain an online prediction algorithm for VARMA processes. However, in order to make this work for EC-VARMA models, we need to adjust the framework in order to handle the complexity of the Error Correction term. For the remainder of this section, we describe the adjustment and term the resulting algorithm EC-VARMA-OGD.}}

\removed{
\postsubsec
\subsection{Approximating an EC-VARMA Process}
\presubsec

Given that an EC-VARMA process starts at some fixed time $t=0$ with fixed initial values, we can write Eq. \ref{eq:ec_varma_form} in a pure EC-VAR form \cite{lutkepohl2006forecasting}:
\begin{equation}
\Delta \vx_t = \Pi^* \vx_{t-1} + \sumi{i}{t-1} \Gamma_i^* \Delta \vx_{t-i} + \bveps_t,\ \ \ t \in \mathbb{N}
\end{equation}
\removed{where $\Pi^*$ and $\Gamma^*_i$ satisfy that
\[
\vTheta(L)^{-1}\left(\mathbf{I}\Delta - \Pi L - \sumi{i}{p-1} \Gamma_i \Delta L^i \right) = \mathbf{I} \Delta - \Pi^* L - \sumi{i}{\infty} \Gamma_i^* \Delta L^i
\]}
This allows us to approximate an EC-VARMA process with an EC-VAR model. \removed{To use EC-VARMA as a DGP in Algorithm \ref{alg:tsp_ogd}, we edit line 5 to be:
\[
\Delta \tilde{\vx}_t = \hat{\Pi} \vx_{t-1} + \sumi{i}{M} \hat{\Gamma}_i \Delta \vx_{t-i}
\]
where $\vgamma = \{\hat{\Pi}, \hat{\Gamma}_1, \ldots, \hat{\Gamma}_M\}$ are the approximating EC-VAR parameters.}
}

\postsubsec
\subsection{Online Prediction for EC-VARMA Models}
\presubsec

We generalize the assumptions U1-U4 to the multivariate setting:
\preitemize
\begin{itemize}
\item[M1)] $\vx_t$ is generated by an EC-VARMA process. The noise sequence $\bveps_t$ of the underlying VARMA process is independent. Also, it satisfies that $\expec[\|\bveps_t\|_2] < M_{\max} < \infty$.
\item[M2)] We overload notation for the vector case and let $\ell_t: \R^{2k} \rightarrow \R$ be a convex loss function with Lipschitz with constant $L > 0$.
\removed{\item[M3)] $\|\vPhi\|_{\max} < C_{\vPhi}$, where $\|\vA\|_{\max} = \max_{ij} \left|a_{ij}\right|$ is an element-wise matrix norm. \textcolor{red}{We don't really need this assumption.. it's only need in the details of the OGD algorithm.}}
\item[M3)] We assume the companion matrix $\vF$ of the MA lag polynomial is diagonalizable. $\lambda_{\max}$ and $\kappa$ are the same as in assumption U4.
\removed{The DGP is invertible. Again, let $\vF$, the companion matrix of the MA lag polynomial, be diagonalizable, i.e. $\vF = \vT \Lambda \vT^{-1}$ where $\vF$ is defined in Eq. \ref{eq:varma_companion_matrix}. $\lambda_{\max}$ and $\kappa$ are the same as in assumption U5.}
\removed{\item[M5)] $\|\vx_t\|_2 < C(t) = o\left(t^{\frac14}\right)$, \textcolor{red}{$\|\Delta \vx_t\|_2 < C_\Delta$}. Note that if the data follows a linear trend, logging it will satisfy these assumptions.}
\end{itemize}
\postitemize

The resulting algorithm is summarized in Algorithm \ref{alg:ec_varma_ogd}, denoted EC-VARMA-OGD. The setup of this algorithm is the same as in Section \ref{sec:online_time_series_prediction_framework}. We overload more notation to generalize Equations \ref{eq:approx_loss} and \ref{eq:transformed_arma_loss}. First, we have $\ell_t^M(\vgamma) :=$
\preeq
\begin{align}
 \ell_t \left( \vx_t,\vx_{t-1} + \hat{\Pi} \vx_{t-1} + \sumi{i}{M} \hat{\Gamma}_i \Delta \vx_{t-i} \right)
\end{align}
\posteq
and $f_t(\Pi, \Gamma, \vTheta) =$
\preeq
\begin{align}
\notag \ell_t \left(\vx_t, \vx_{t-1} + \Pi \vx_{t-1} + \sumi{i}{p-1} \Gamma_i \Delta \vx_{t-i} + \sumi{i}{q} \vTheta_i \bveps_{t-i} \right)
\end{align}
\posteq
where $\vgamma = \{\hat{\Pi}, \hat{\Gamma}_1, \ldots, \hat{\Gamma}_M\}$ are the approximating EC-VAR parameters. The regret as defined in Eq. (\ref{eq:univariate_extended_regret}) can be generalized to
\preeq
\begin{equation} \label{eq:ec_varma_ogd_extended_regret}
\text{Regret} = \sumi{t}{T} \ell_t^M (\vgamma_t) - \min_{\Pi, \Gamma, \vTheta \in \mathcal{K}} \sumi{t}{T} \expec [f_t(\Pi, \Gamma, \vTheta)]
\end{equation}
\posteq
where $\mathcal{K}$ is the set of invertible EC-VARMA processes.

To encourage $\hat{\Pi}$ to be low rank, we project it onto $\mathcal{B}(\cdot,\rho)$, which is the nuclear norm ball of radius $\rho$. This involves projecting the singular values of $\hat{\Pi}$ onto an $\ell_1$-ball and can be efficiently done \citep{duchi2008efficient}. In our framework, this is handled by letting the convex set $\mathcal{E}$ be described as $\{\vgamma : \|\hat{\Pi}\|_* \leq \rho, \|\hat{\Gamma}_i\|_{\max} \leq 1,\ i = 1, \ldots, M \}$ and plugging it into OGD where projections are made at each iteration. For convenience of notation, let $\mathcal{E}_\Gamma = \{ \hat{\Gamma} : \|\hat{\Gamma}_i\|_{\max} \leq 1, i = 1, \ldots M\}$. As in Section \ref{sec:online_time_series_prediction_framework}, $\mathcal{E}$ should be chosen to be large enough to encompass a valid approximation to the true DGP. \removed{In practice one will choose $\rho$ and $\mathcal{E}_\Gamma$ to be something simple.}

\begin{algorithm}[t]
\caption{EC-VARMA-OGD}
\label{alg:ec_varma_ogd}
\begin{algorithmic}[1]
\REQUIRE DGP parameters $p,q$. Horizon $T$. Learning rate $\eta$. Data: $\{\vx_t\}$.
\STATE Set $M = \log_{\lambda_{\max}} \left(\left( 2\kappa TLM_{\max}\sqrt{q} \right)^{-1} \right) + p $
\STATE Choose $\vgamma^{(1)} \in \mathcal{E}$ arbitrarily.
\FOR {$t=1$ to $T$}
\STATE Predict $\tilde{\vx}_{t} = \vx_{t-1} + \hat{\Pi} \vx_{t-1} + \sumi{i}{M} \hat{\Gamma}_i \Delta \vx_{t-i}$
\STATE Observe $\vx_t$ and receive loss $\ell_t^M\left(\vgamma^{(t)}\right)$
\STATE \removed{Set }$\hat{\Gamma}^{(t+1)}_i = \Pi_{\mathcal{E}_\Gamma}\left( \hat{\Gamma}^{(t)}_i - \eta \nabla_{\Gamma_i} \ell_t^M\left(\vgamma^{(t)}\right) \right)$\removed{, for all $i$}
\STATE \removed{Set }$\hat{\Pi}^{(t+1)} = \Pi_{\mathcal{B}(*, \rho)}\left( \hat{\Pi}^{(t)} - \eta_t \nabla_\Pi \ell_t^M\left(\vgamma^{(t)}\right) \right)$
\ENDFOR
\end{algorithmic}
\end{algorithm}

We present the following regret bound:

\begin{theorem} \label{thm:ec_varma_ogd} Let $\eta = \frac{D}{G(T)\sqrt{T}}$. Then for any data sequence $\{\vx_t\}_{t=1}^T$ that satisfies assumptions M1-M3, Algorithm \ref{alg:ec_varma_ogd} generates a sequence $\{\vgamma_t\}$ in which
\preeq
\begin{equation*}
\textrm{Regret} \removed{\sumi{t}{T} \ell_t^M (\vgamma_t) - \min_{\Pi, \Gamma, \vTheta \in \mathcal{K}} \sumi{t}{T} \expec [f_t(\Pi, \Gamma, \vTheta)]} = O\left( DG(T) \sqrt{T} \right)
\end{equation*}
\posteq
\end{theorem}
For the remainder of the section, we assume that $\ell_t$ is the squared loss and $\|\vx_t\|_2 < C(t) = O( \log t )$.

\emph{Remark 1:} With the above assumptions, the resulting regret bound of EC-VARMA-OGD is $O\left(k^2 M^2 \log^2(T) \sqrt{T} \right)$.

\emph{Remark 2:} By setting $\rho = 0$ and using $\vx_t$ in place of $\Delta \vx_t$ (i.e. not differencing) in Algorithm \ref{alg:ec_varma_ogd}, we effectively use a VARMA process as the DGP and achieve an \textit{equivalent regret bound} as in the previous remark. Denote this adaptation as VARMA-OGD. However, if the DGP is EC-VARMA, we expect this to empirically perform worse than EC-VARMA-OGD since the latter exploits a valid transformation of the data.

\emph{Remark 3:} Assume that the DGP is an EC-VARMA process and $\rho = o(1 / \log^2 (T) )$. Then the regret bound obtained is $O\left(k^2 M^2 \sqrt{T}\right)$. In Section \ref{sec:empirical_results}, we find that this choice of $\rho$ works well empirically.

\removed{For the squared loss, assuming a EC-VARMA DGP, $C(T) = O(\log(T))$ and $\rho = o\left( 1/\log^2 (T) \right)$ then the regret bound we obtain is $O \left(\sqrt{T} \right)$. As discussed before, the $\log$-transformation is very common in applications of time series methods and therefore the bound on $C(T)$ is justified.}

\removed{
\begin{table}[t]
\caption{Regret Bounds for Different Assumed DGPs when true DGP is SARIMA}
\label{table:multivariate_regret_comparison}
\centering
\begin{tabular}{|c|c|}
\hline
Assumed DGP & Regret Bound \\ \hline
VARMA & $O\left( C^2(T) k^2 M^2 \sqrt{T} \right)$ \\ \hline
EC-VARIMA & $O\left( C^2(T) k^2 M^2 \sqrt{T} \right)$ \\ \hline
\end{tabular}
\end{table}

We consider when $\ell_t(\vx, \vy) = \frac12 \|\vx - \vy\|_2^2$ and the true DGP is an EC-VARMA model. By plugging in assumed DGP's of VARMA and EC-VARMA, we achieve regret bounds as shown in Table \ref{table:multivariate_regret_comparison}. This table considers the theoretical aspects of accounting for non-stationarities in the multivariate domain. Unintuitively, we see that accounting for the cointegration relationship does not provide us with a theoretically faster convergence, which is due to the error correction term. However, we should hope that $\|\hat{\Pi} \vx_{t-1}\|_2$ she be small after some $t$ large enough, which says that we have found a cointegrating relationship. This would give us faster empirical convergence. Also, limiting $\rho$ should bound the error correction term, which leads us to this remark:
\begin{remark}
Assume the true DGP is an EC-VARMA model. Let $\rho = o\left( \frac{1}{\log^2(T)} \right)$ and $C(T) = O(\log(T))$. Note that if the data follows a linear trend, logging the data will achieve the bound. Then the resulting regret bound of EC-VARMA-OGD is
\[
O \left( C_\Delta k^2M^2\sqrt{T} \right)
\]
\end{remark}
}

\postsec
\section{NonSTOP}
\label{sec:unknown_transformation}
\presec

Algorithms \ref{alg:tsp_ogd} and \ref{alg:ec_varma_ogd} assume that the appropriate transformation is known apriori. Typically, statistical tests are used to determine the degree of differencing on a fixed dataset (e.g. \cite{EJT1996}) and these usually come with assumptions and sample size requirements. In the online setting, these requirements are not realistic and an ideal method must adapt to the incoming data from a possibly time dependent sequence of transformations. We approach this problem by using the online learning with experts (OLE) setup wherein each expert corresponds to a specific transformation (including the identity transform). Specifically, we adapt the (randomized) weighted majority algorithm \citep{shalev2011online} as a meta-algorithm to select a transformation at each time step.\removed{ and allow for potentially unbounded loss functions.}

More precisely, let $\mathcal{M}$ be the set of experts we consider. The set of experts can either be instantiations of Algorithm \ref{alg:tsp_ogd} or \ref{alg:ec_varma_ogd}. For example, in the univariate setting, we could have $\mathcal{M} = \{\text{ARMA-OGD}, \text{ARIMA-OGD}, \text{SARIMA-OGD} \}$ with both $d$ and $\tilde{D}$ set to $1$, and in the multivariate setting we can have $\mathcal{M} = \{\text{VARMA-OGD}, \text{EC-VARMA-OGD} \}$. We assume that the seasonal period $s$ is known. 

\removed{Every expert is an instantiation of Algorithm \ref{alg:tsp_ogd} using a transformation such that $ d \in \{0, 1, \ldots d_{\textrm{max}}\}, \tilde{D} \in \{0, 1, \ldots \tilde{D}_{\textrm{max}}\}$, where $d_{\textrm{max}}, \tilde{D}_{\textrm{max}}$ are the largest differencing and seasonal differencing order we consider, respectively. For example, $d = \tilde{D} = 0$ results in ARMA-OGD. We assume that the seasonal period $s$ is given.  In practice, we rarely deal with $d_{\textrm{max}} > 1$ or $\tilde{D}_{\textrm{max}} > 1$, thus choose $d_{\textrm{max}} = \tilde{D}_{\textrm{max}} = 1$ in Section \ref{sec:empirical_results}.}

\begin{algorithm}[t]
\caption{NonSTOP}
\label{alg:tsp_ogd_unknown_transformation}
\begin{algorithmic}[1]
\REQUIRE DGP parameters $l_a, l_m$. Horizon $T$. Data: $\{x_t\}$. Models $\mathcal{M}$. Window size $k$.
\STATE Set $\eta = \sqrt{\frac{\log |\mathcal{M}| }{T}}$
\STATE Initialize $\boldsymbol{w}_1 = \colvec{1 & \ldots & 1}$
\FOR {$t=1$ to $T$}
\STATE Set $b_t = \max_{\tau \in \{t-k, \ldots. t\}, h \in \mathcal{M}} \ell_{\tau}(h), W_t = \sum_h w_t(h)$
\STATE Predict using $h_t \in \mathcal{M}$, where $h_t$ is chosen using probability distribution $\frac{\boldsymbol{w}_t}{W_t}$
\STATE For each model in $\mathcal{M}$, run the update according to Algorithm \ref{alg:tsp_ogd}.
\STATE Update ${w}_{t+1}[h] = {w}_t[h] (1 - \eta)^{\frac{\ell_t(h)}{b_t}}$ for all $h \in \mathcal{M}$
\ENDFOR
\end{algorithmic}
\end{algorithm}

The resulting algorithm refered to as NonSTOP is summarized in Algorithm \ref{alg:tsp_ogd_unknown_transformation}. In each round, the online meta-algorithm randomly selects a prediction from one of its experts. After receiving the loss, it then updates its view about its experts, while the experts themselves are adapting to the data. We scale the loss function with a sliding window maximum such that the losses stay bounded. Since $D, G(T)$, and $\ell_t^M\left(\vgamma^{(t)}\right)$ as shown in Algorithm \ref{alg:tsp_ogd} and \ref{alg:ec_varma_ogd} are now dependent on the specific transformation, we denote this as $D_h, G_h(T), \ell_{t,h}^M\left(\vgamma_h^{(t)}\right)$ for a model $h \in \mathcal{M}$. Define Regret = $\sumi{t}{T} \expec \left[ \ell_t(h_t) \right] - \min_{\valpha, \vbeta \in \mathcal{K}} \sumi{t}{T} \expec\left[ f_t(\valpha, \vbeta) \right]$, where $\ell_t(h) := \ell_{t,h}^M \left(\vgamma^{(t)}_h\right)$. With these definitions in hand, we give the following theorem:

\begin{theorem} \label{thm:tsp_ogd_unknown_transformation} Define $B_T := \max_{\tau \in \{1, \ldots, t\}, h \in \mathcal{M}} \ell_{\tau}(h)$. Then Algorithm \ref{alg:tsp_ogd_unknown_transformation} plays a sequence of predictions that satisfies
\preeq
\begin{equation*}
\textrm{Regret} = O \left(  \max\left\{ B_T, D_*G_*(T)\right\} \sqrt{T} \right)
\end{equation*}
\posteq
where $D_* = \max_h D_h,\ G_*(T) = \max_h G_h(T)$. \removed{Note that this bound is the same as if you used $B_T$ in place of $b_t$.}

\end{theorem}
\begin{remark} \label{remark:least_squares_loss}
When using least squares loss, $B_T = O(G_*(T))$ and the regret bound defaults to $O \left( D_* G_*(T) \sqrt{T} \right)$.
\end{remark}

\postsec
\section{Empirical Results}
\label{sec:empirical_results}
\presec

In this section, we show empirically the effectiveness of methods described in Sections \ref{sec:online_time_series_prediction_framework}, \ref{sec:error_corrected_varma_models}, and \ref{sec:unknown_transformation} on synthetic and real datasets. In each scenario, we use squared loss and plot the log average squared loss vs. iteration. For all experiments, we set $\mathcal{E} = \{\vgamma : \|\vgamma\|_{\max} \leq 1\}$, initialize all parameters to 0, and set the sliding window length $k = 10$. For all real world datasets, we log transform the time series. Plots of these datasets can be found in the Supplement.

\begin{figure*}[t]
\centering
\subfloat[Synthetic data] {\label{fig:results_synthetic}\includegraphics[width=0.25\linewidth]{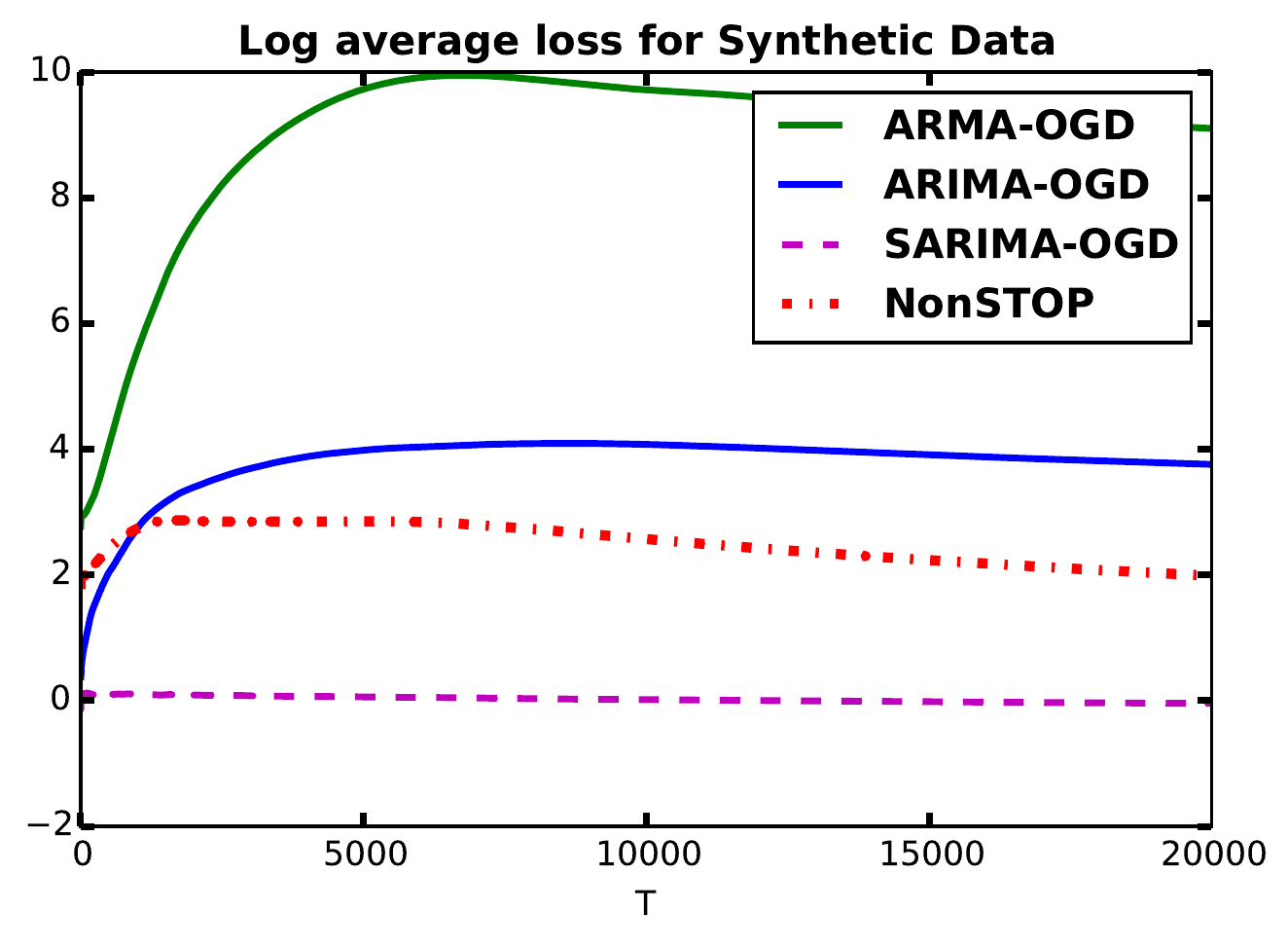}}
\subfloat[Synthetic switching data] {\label{fig:results_synthetic_switch}\includegraphics[width=0.25\linewidth]{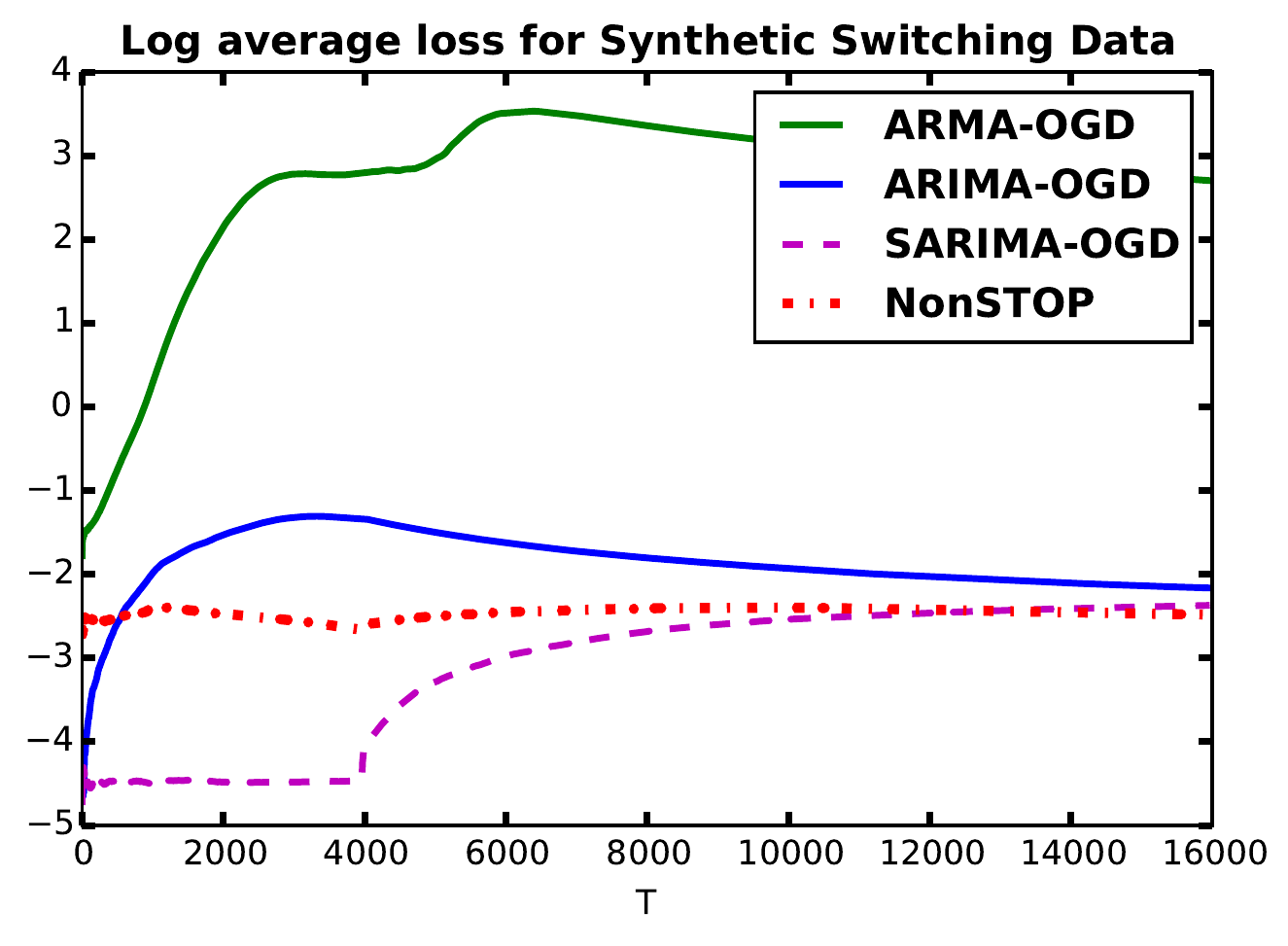}}
\subfloat[Turkey electricity demand]{\label{fig:results_turkey_elec}\includegraphics[width=0.25\linewidth]{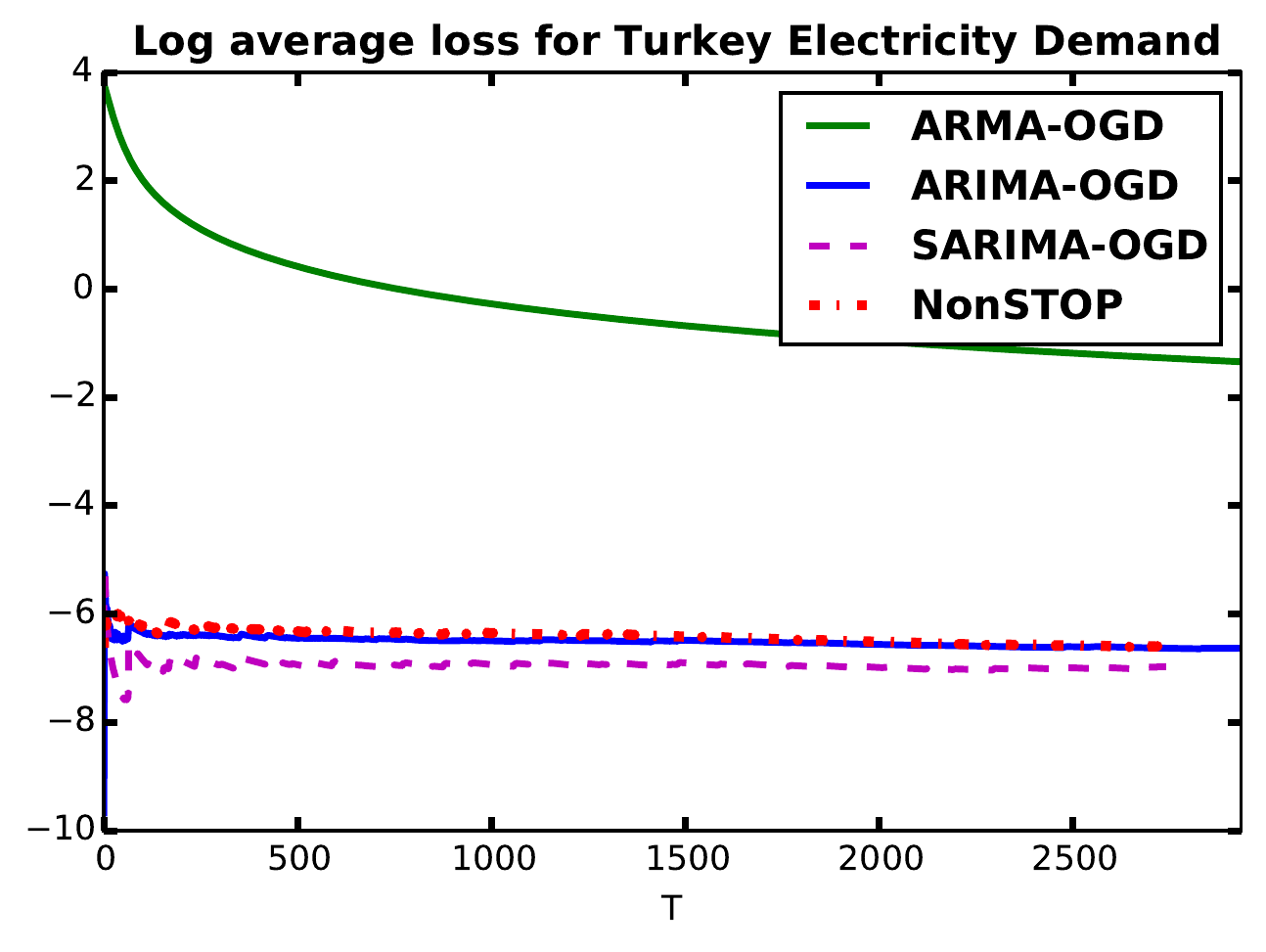}}

\medskip
\subfloat[Births in Quebec]{\label{fig:results_quebec_births}
\includegraphics[width=0.25\linewidth]{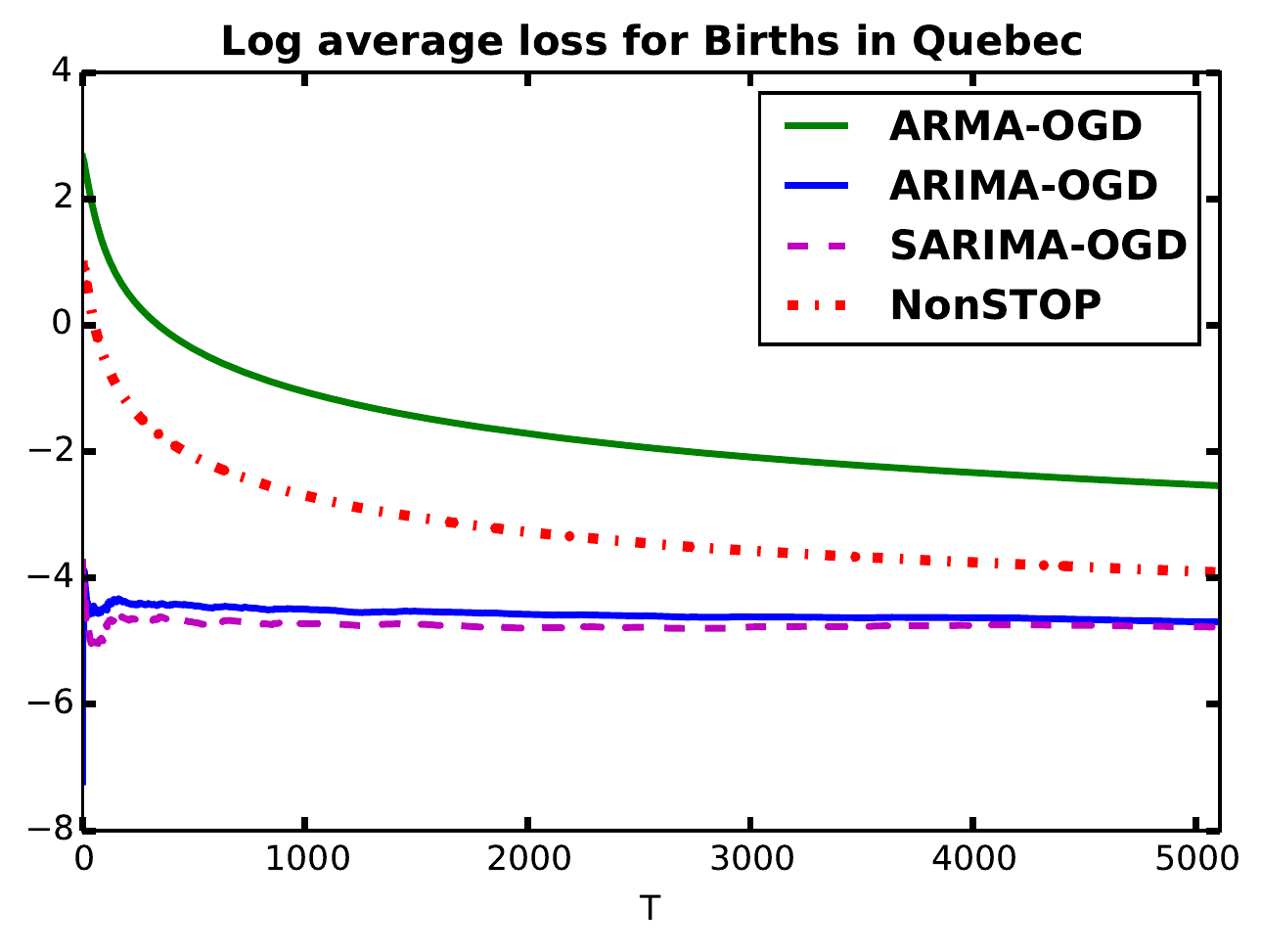}}
\subfloat[Saugeen River flow] {\label{fig:daily_river_flows}\includegraphics[width=0.25\linewidth]{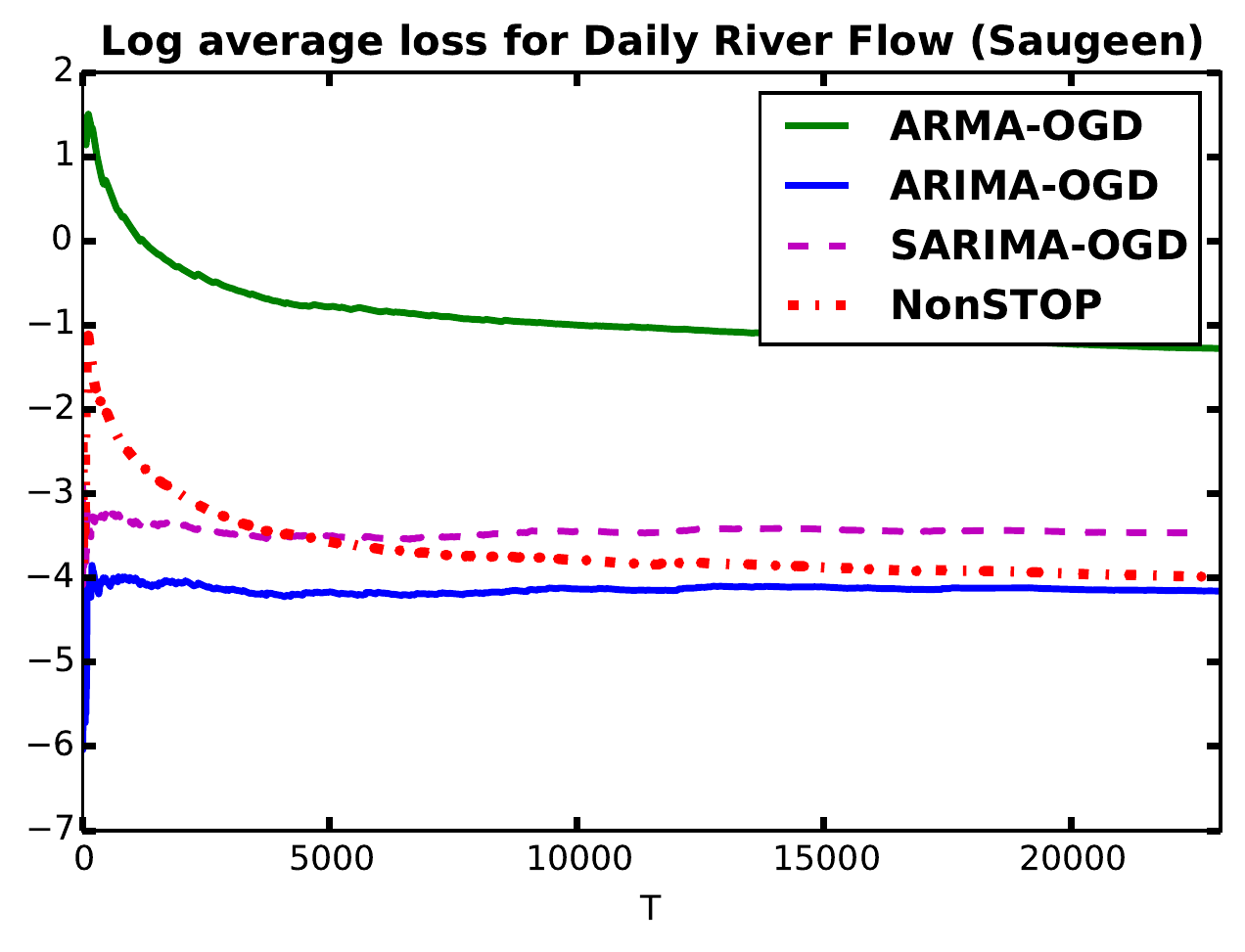}}
\subfloat[Stock data]{\label{fig:stock_data}
\includegraphics[width=0.25\linewidth]{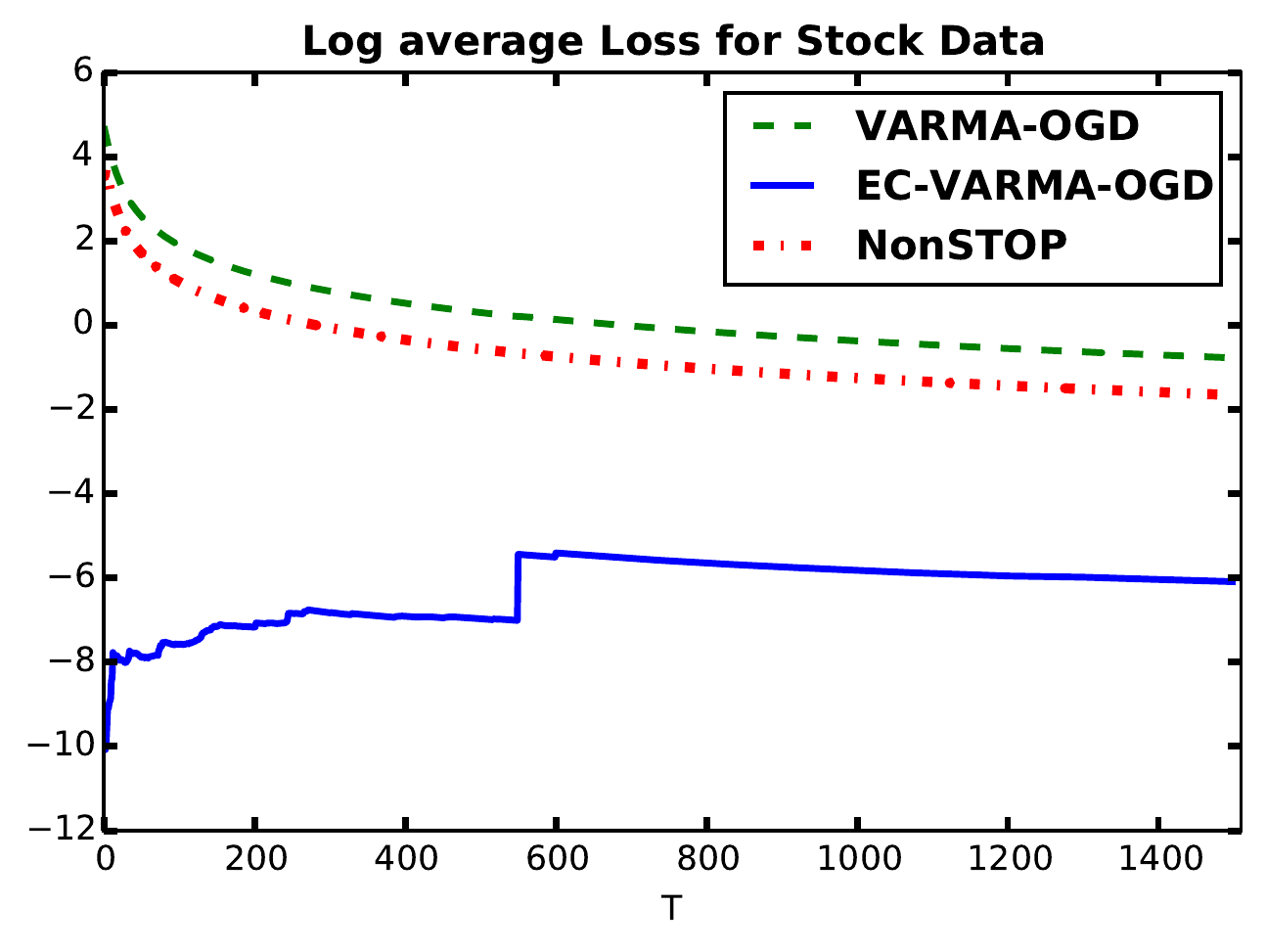}}
\subfloat[Google flu data]{\label{fig:flu_data}\includegraphics[width=0.25\linewidth]{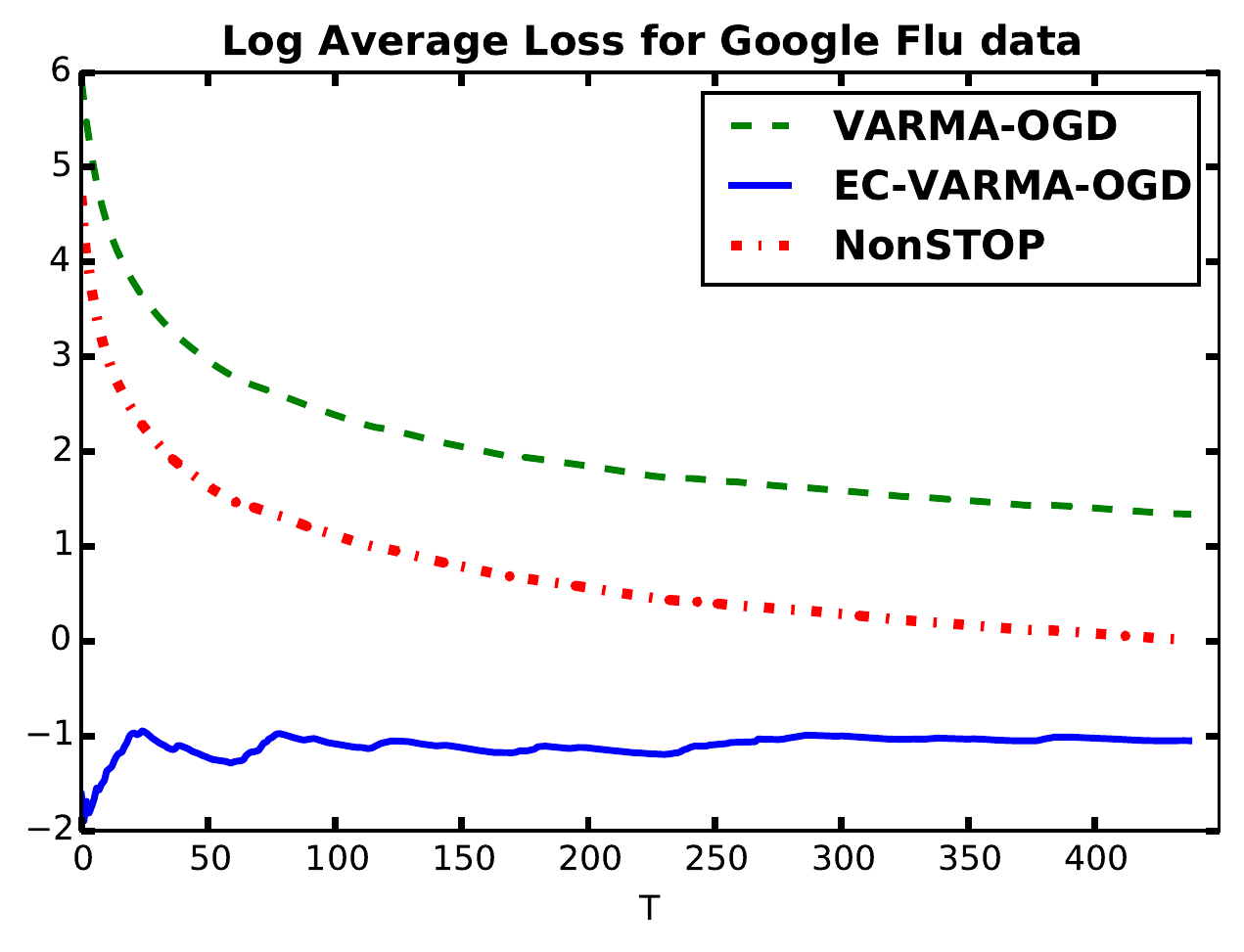}}

\caption{Empirical results for synthetic and real datasets}
\end{figure*}

\postsubsec
\subsection{Univariate Setting}
\presubsec

To illustrate that applying transformations accounting for appropriate nonstationarities results in superior empirical performance, we compare the algorithms given in Table \ref{table:univariate_regret_comparison}. We also include a comparison to NonSTOP to showcase its efficacy in a fully online setting. For each dataset, we assume the seasonal differencing order $\tilde{D}=1$. We set $M= 2s$ and $d=1$ for each dataset.

We first simulate 20 synthetic time series with $T = 20000$ from the following SARIMA model:
\preeq
\begin{align}
\Delta \Delta_{12} x_t = (1-0.95L)(1-0.4L^{12})\epsilon_{t}
\end{align}
\posteq
We run the algorithms on each generated time series and average the log average loss in Figure \ref{fig:results_synthetic}. As expected, SARIMA-OGD outperforms ARMA-OGD and ARIMA-OGD, converging quickly as it accounts for the appropriate nonstationarities. This behaviour is consistent with our hypothesis that in the absence of an appropriate transformation, existing methods will underperform. NonSTOP gradually adapts and learns to heavily weight the correct transformation (expert) and outperforms ARMA-OGD and ARIMA-OGD. \removed{Note that the initial bump in the convergence of ARMA-OGD is due to the fact that we are plotting (log) average loss and there is an initial period where the average loss grows faster than $T$, and then it grows slower.}

To showcase the adaptability of NonSTOP, we simulate 20 synthetic time series from Eq. (16) for $4000$ timesteps from a SARIMA model, and then simulate data from an ARIMA model for another $12000$ timesteps. Results are shown in Figure \ref{fig:results_synthetic_switch}. NonSTOP learns to weight SARIMA-OGD initially, but quickly adapts at $t=4000$ to weight ARIMA-OGD. In fact, at the end of the run, it actually outperforms all experts, showing the power of this online adaptable algorithm.

Next, we consider a dataset that contains daily electricity demand in Turkey\removed{ from January 1, 2000, to December 31, 2008}. The seasonality in this dataset is biannual\removed{ (rounded down to $s=182$ days)}. The results of running the algorithms are shown in Figure \ref{fig:results_turkey_elec}. Again, SARIMA-OGD accounts for the appropriate nonstationarities and performs the best. ARMA-OGD suffers severely due to not accounting for any nonstationarity. As such, NonSTOP quickly finds that ARMA-OGD is not a reliable expert and performs well in comparison to the other experts. For the daily recorded births in Quebec\removed{ from Jan. 01, 1977 to Dec. 31, 1990}, there is a weekly seasonality pattern with $s = 7$. Figure \ref{fig:results_quebec_births} reveals that the results here are similar to previous datasets. Because NonSTOP starts with an equal weight for each expert, it pays a large penalty for selecting ARMA-OGD in initial iterations. However, it approaches the performance of the other algorithms as it learns to heavily weight the correct transformation.

Lastly, we consider a dataset consisting of daily river flow values from the Saugeen River\removed{ from Jan 01, 1915 to Dec 31, 1979}. This data seems to exhibit a yearly ($s=365$) seasonality pattern (see supplement). The results are plotted in Figure \ref{fig:daily_river_flows}. While accounting for any nonstationarity improves performance, accounting for seasonality actually hurts the performance compared to accounting for only trend\removed{ as SARIMA-OGD is outperformed by ARIMA-OGD}. In our experience, ARIMA can sometimes outperform SARIMA even on seemingly seasonal data. Despite this, NonSTOP learns to weight ARIMA-OGD and quickly approaches the best performance. This showcases the efficacy of the NonSTOP algorithm in a fully online setting.

\removed{Next, we consider two real world datasets. The first is a dataset that contains daily electricity demand in Turkey from January 1, 2000, to December 31, 2008, which we got from \cite{}. The seasonality in this dataset is bi-annual (rounded down to $s=182$ days). This dataset exhibits seasonality and an upwards trend, making it susceptible to being well-modeled by a SARIMA process. The second dataset contains daily recorded births in Quebec from Jan. 01, 1977 to Dec. 31, 1990. There is a weekly seasonality pattern, thus $s = 7$. The results are in Figure \ref{fig:results_turkey_elec} and \ref{fig:daily_river_flows}. In both settings, accounting for any nonstationary results in faster convergence compared to the non-adjusted transform. Accounting for seasonality as well as the trend in the transform results in slightly faster convergence than only accounting for the trend.

Lastly, we consider a dataset consisting of daily river flow values from the Saugeen River from Jan 01, 1915 to Dec 31, 1979. This data exhibits a yearly ($s=365$) seasonality pattern. The results are plotted in Figure \ref{fig:daily_river_flows}. Note that while accounting for any non-stationarity still improves convergence, accounting for the seasonal non-stationarity actually hurts the performance compared to accounting for only the trend. We hypothesize that the data is not well fit by a SARIMA model, and that we are instead forcing seasonality into the model. Note that ARIMA processes can model non-stationarities such as seasonality as well (and so can ARMA), thus a seasonal transform is not guaranteed to always improve convergence, even when the data does exhibit such patterns.}

\postsubsec
\subsection{Multivariate Setting}
\presubsec

\removed{
\begin{figure*}[t]
\centering
\subfloat[Stock data]{\label{fig:stock_data}
\includegraphics[width=0.32\linewidth]{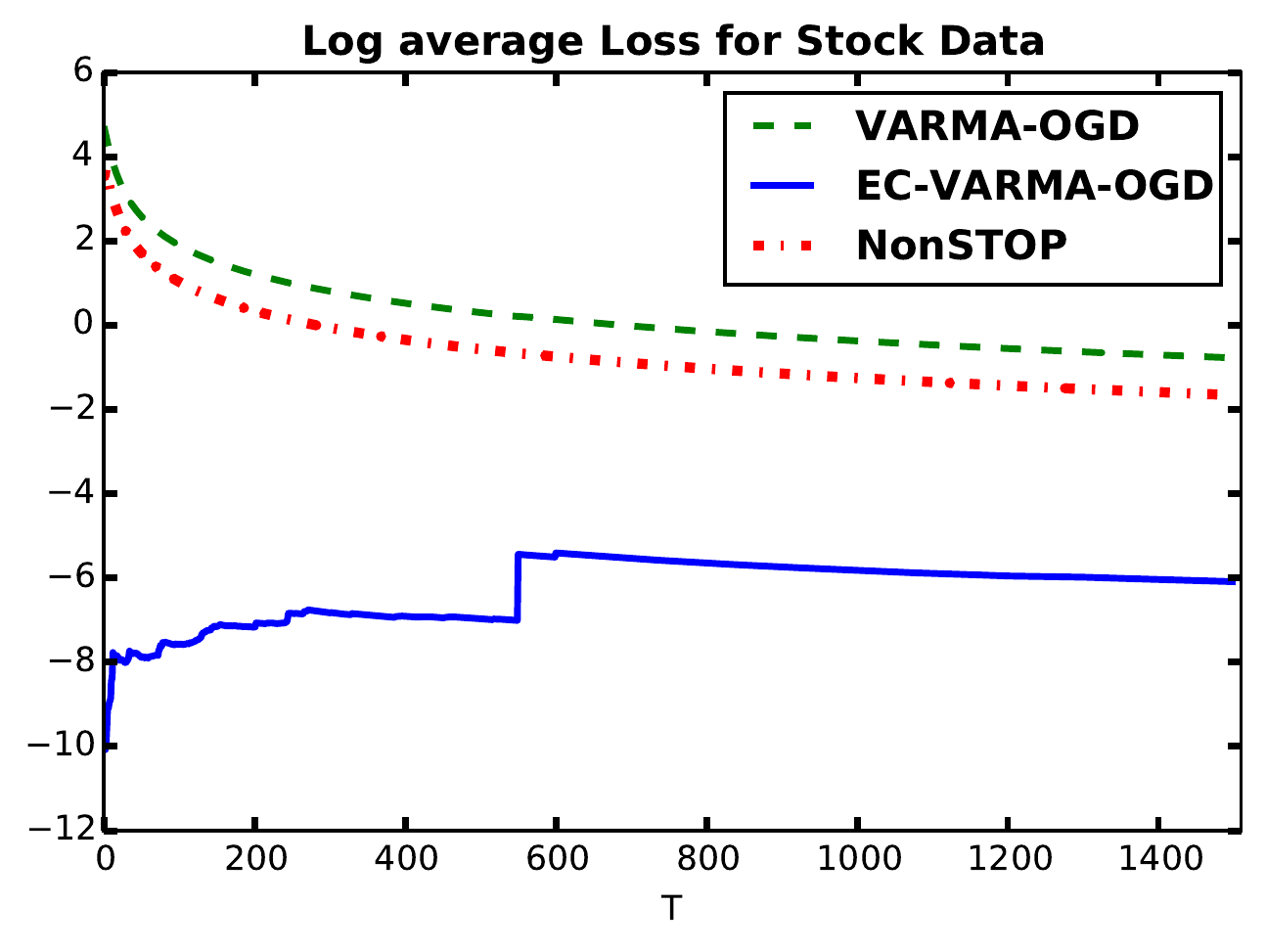}}
\subfloat[Google flu data]{\label{fig:flu_data}\includegraphics[width=0.32\linewidth]{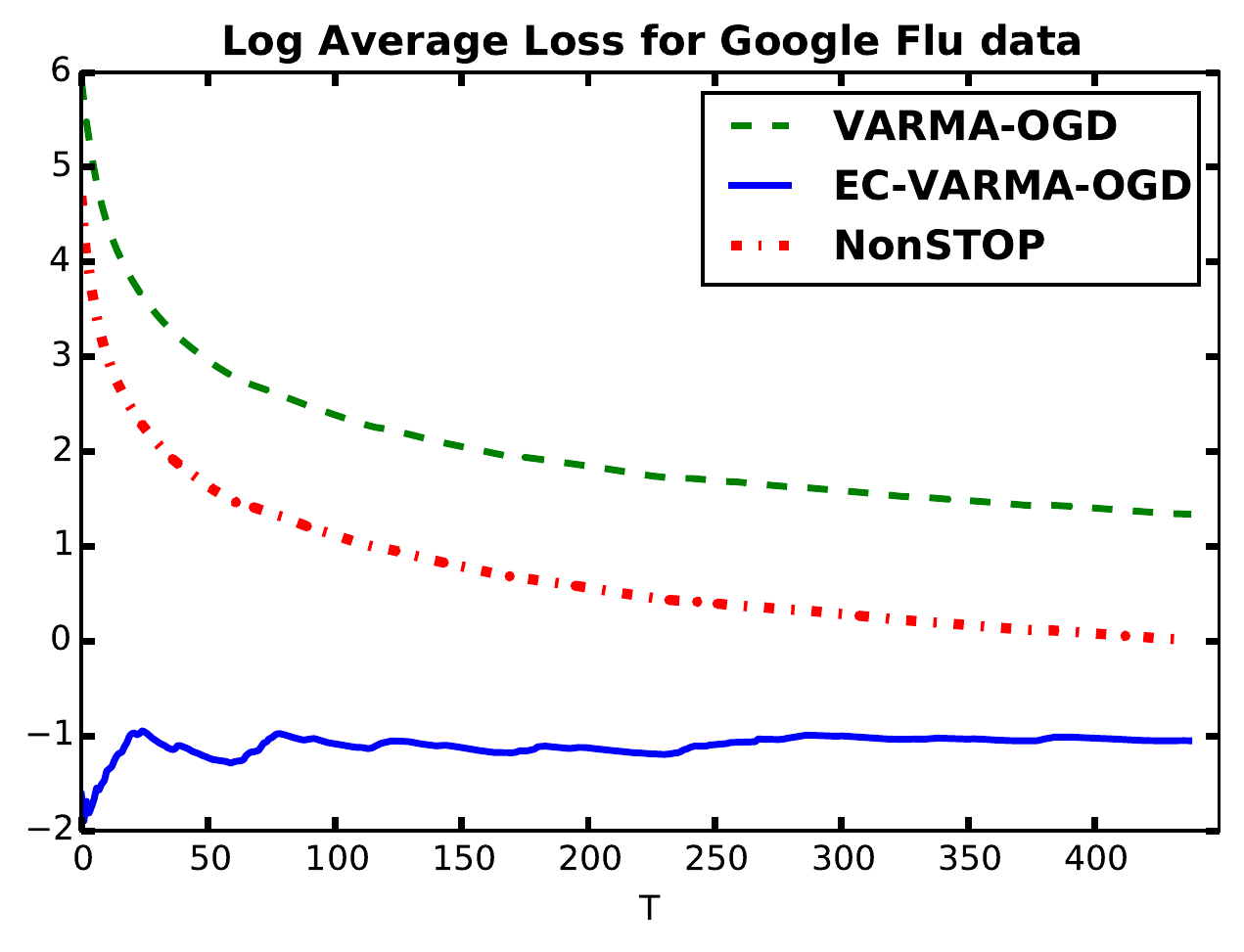}}

\caption{Empirical results for multivariate datasets}
\end{figure*}
}

\removed{In the multivariate setting we show empirically that accounting for cointegration results in faster convergence. We look at the results of running EC-VARMA-OGD as described in Algorithm \ref{alg:ec_varma_ogd} vs. VARMA-OGD on two real datasets. }

We collected 7 time series of stock prices from Yahoo Finance\removed{ (\url{http://finance.yahoo.com/})} of large technology companies, and also included the S\&P500 index. By including the S\&P500, which is essentially an weighted average of 500 company stock prices, we have partially introduced cointegration into the time series. We set $M=10, \rho=0.5$ and ran all algorithms with the resulting plots in Figure \ref{fig:stock_data}. Accounting for cointegration results in considerably stronger performance. There is a bump in the convergence plot due to a spike in the data (see Supplement). We also evaluated the algorithms on the Google Flu dataset\removed{ (\url{https://www.google.com/publicdata/explore/}) which contains influenza rates of $28$ countries}.  There are two distinct seasonality patterns: the northern hemisphere countries have flu incidents that peak in one part of the year while the southern hemisphere countries have flu incidents that peak in the other part of the year. Thus, it makes sense to believe that the time series exhibit cointegration. This dataset exhibits yearly seasonality, thus we set $M = 60$ to be larger than one seasonal period. We choose $\rho = 0.5$ and plot the results are given in Figure \ref{fig:flu_data}. Again, adjusting for the cointegration dramatically increases predictive performance. 

On both datasets, NonSTOP pays a penalty for selecting VARMA-OGD in the initial iterations before learning to heavily weight EC-VARMA-OGD. Note that NonSTOP outperforms VARMA-OGD by at least a factor of 3 on the original scale for both datasets.

\postsec
\section{Conclusions and Future Work}
\presec

We presented general online time series methods that account for nonstationary artifacts in both univariate and multivariate data. If such artifacts are known in advance, we demonstrate that these transformations lead to superior theoretical and empirical performance. Speculating that accounting for nonstationarities reduces correlation in the data, we presented a data dependent regret bound for FTL in the case of squared loss. In the case that the artifacts are unknown, we incorporate a finite set of possible transformations into a OLE framework called NonSTOP that can learn to appropriately weight the correct transformation. We provided empirical results showing the efficacy of our proposed methods. In future work, we plan to explore extensions that hold for more complicated models including long memory models such as ARFIMA.

\postsec

\label{Bibliography}
\bibliography{opt2}

\appendix

\section{More Background: Companion Matrix}

If the MA lag polynomial of a SARIMA process has all of its roots outside of the complex unit circle, then the SARIMA process is defined as invertible. Let $\beta_i$ be the scalar coefficients of the MA lag polynomial (recall that this is $\theta(L)\Theta(L^s)$). Invertibility is equivalent to saying that the companion matrix
\begin{equation*} \label{eq:companion_matrix}
\vF = \colvec{-\beta_1 & -\beta_2 & \ldots & \ldots & -\beta_{l_m} \\ 1 & 0 & \ldots & \ldots & 0 \\ 0 & 1 & 0 & \ldots & \vdots \\ \vdots & 0 & \ddots & \ddots & \vdots \\ 0 & \vdots & \vdots & 1 & 0}
\end{equation*}
has eigenvalues less than 1 in magnitude. If this is the case, then the underlying ARMA process $\Delta^d \Delta^{\tilde{D}}_s x_t$ can be written as an AR$(\infty)$ process and can be approximated by a finite truncated AR process. 

In the multivariate setting, the requirements for invertibility are very similar to the univariate case. For EC-VARMA (and VARMA) processes, we require that $\det\left(\vTheta(L) \right)$ must have all of its roots outside of the complex unit circle. Again, this is equivalent to saying that the companion matrix has eigenvalues less than 1 in magnitude \citep{lutkepohl2005new, tsay2013multiple}. If the process is invertible, then it can be rewritten as a VAR$(\infty)$ process.

\section{Proof of Theorem \ref{thm:tsp_ogd}}

We give a proof similar to \citet{AnavaHazan} and \citet{liu2016online} using our transformation notation, and with the more natural and relaxed assumption of invertibility of the MA process.

\begin{proof}

\textbf{Step 1}: Assume that $\zeta(\tilde{x}_t)$ is a linear function such as the ones given in Table \ref{table:dgp}. Then $\{\ell_t^M\}$ are convex loss functions, and we may invoke \citep{Zinkevich2003OnlineCP} with a fixed step size $\eta = \frac{D}{G(T)\sqrt{T}}$:
\[
\sumi{t}{T} \ell_t^M(\vgamma_t) - \min_{\vgamma} \sumi{t}{T} \ell_t^M(\vgamma) = O\left( DG(T) \sqrt{T} \right)
\]
Note that the proof in \citep{Zinkevich2003OnlineCP} uses a constant upper bound $G$ on the gradients. Since we assume $G(T)$ is a monotonically increasing function, the proof in \citep{Zinkevich2003OnlineCP} follows through straightforwardly.

\textbf{Step 2}: Let $\valpha, \vbeta$ denote the parameters of the underlying ARMA$(l_a,l_m)$ process. We define a few things:
\bal
\tau\left( x_t^\infty(\valpha, \vbeta) \right) &= \sumi{i}{l_a} \alpha_i \tau\left( x_{t-i}\right) + \sumi{i}{l_m} \beta_i \left( \tau\left( x_{t-i}\right) - \tau\left( x_{t-i}^\infty (\valpha, \vbeta) \right) \right)\\
x_t^\infty(\valpha, \vbeta) &=\zeta\left( \tau\left( x_t^\infty(\valpha, \vbeta) \right) \right)
\eal
with initial condition $\tau\left( x_t^\infty (\valpha, \vbeta) \right) = \tau\left( x_t \right)$ for $t < 0$. For convenience, assume that we have fixed data $x_{0}, \ldots, x_{-h}$ so that $\tau(x_0), \ldots, \tau(x_{-l_a})$ exists. Denote
\[
f_t^\infty(\valpha, \vbeta) = \ell_t\left(x_t, x_t^\infty(\valpha, \vbeta) \right)
\]
With this definition, we can write $\tau\left( x_t^\infty(\valpha, \vbeta) \right) = \sumi{i}{t+l_a} c_i(\valpha, \vbeta) \tau\left( x_{t-i} \right)$, i.e. as a growing AR process. Next, we define
\bal
\tau\left( x_t^m(\valpha, \vbeta) \right) &= \sumi{i}{l_a} \alpha_i \tau\left(x_{t-i}\right) + \sumi{i}{l_m} \beta_i \left( \tau\left( x_{t-i}\right) - \tau\left( x_{t-i}^{m-i} (\valpha, \vbeta) \right) \right)\\
x_t^m(\valpha, \vbeta) &= \zeta\left(\tau\left( x_t^m(\valpha, \vbeta) \right) \right)
\eal
with initial condition $\tau\left( x_t^m (\valpha, \vbeta) \right) = \tau\left( x_t\right)$ for $m < 0$. We relate $M$ and $m$ with this relation: $M = m + l_a$. With this definition, we can write $\tau\left( x_t^m(\valpha, \vbeta)\right) = \sumi{i}{M} \tilde{c}_i(\valpha, \vbeta) \tau\left( x_{t-i} \right)$, i.e. as a fixed length AR process. Denote
\[
f_t^m(\valpha, \vbeta) = \ell_t\left(x_t, x_t^m(\valpha, \vbeta) \right)
\]
Let $(\valpha^*, \vbeta^*) = \argmin_{\valpha, \vbeta \in \mathcal{K}} \sumi{t}{T} \expec \left[ f_t( \valpha, \vbeta )\right]$. Recall that the only random part of the expectation is $\veps_t$. $x_t$ is fixed in this quantity.

Lemma \ref{lemma:ltm_to_ftm} gives us that
\[
\min_{\vgamma} \sumi{t}{T} \ell_t^M({\vgamma}) \leq \sumi{t}{T} f_t^m(\valpha^*, \vbeta^*)
\]
Lemma \ref{lemma:ftm_to_ftinf} says that choosing $m = \log_{\lambda_{\max}}\left(\left( 2\kappa TLM_{\max}\sqrt{l_m}\right)^{-1}\right)$ results in
\[
\left | \sumi{t}{T} \expec[f_t^m(\valpha^*, \vbeta^*)] - \sumi{t}{T}\expec[f_t^\infty(\valpha^*, \vbeta^*)] \right| = O(1)
\]
Lemma \ref{lemma:ftinf_to_ft} gives us that
\[
\left | \sumi{t}{T} \expec[f_t^\infty(\valpha^*, \vbeta^*)] - \sumi{t}{T}\expec[f_t(\valpha^*, \vbeta^*)] \right| = O(1)
\]
Chaining all of these gives us the final result:
\[
\sumi{t}{T} \ell_t^m(\vgamma_t) - \min_{\valpha, \vbeta \in \mathcal{K}} \sumi{t}{T} \expec[f_t(\valpha, \vbeta)] = O\left(DG(T) \sqrt{T}\right)
\]

\end{proof}

\begin{lemma} \label{lemma:ltm_to_ftm}
For all $m$ and $\{x_t\}$ that satisfies the assumptions U1-U4, we have that
\[
\min_{\vgamma} \sumi{t}{T} \ell_t^m({\vgamma}) \leq \sumi{t}{T} f_t^m(\valpha^*, \vbeta^*)
\]
\end{lemma}
\begin{proof}
We simply set $\gamma_i^* = \tilde{c}_i(\valpha^*, \vbeta^*)$ and get $\sumi{t}{T} \ell_t^m({\vgamma}^*) = \sumi{t}{T} f_t^m(\valpha^*, \vbeta^*)$. Thus, the minimum holds trivially. Note that we assume ${\vgamma}^* \in \mathcal{E}.$
\end{proof}

\begin{lemma} \label{lemma:ftinf_to_ft}
For any data sequence $\{x_t\}$ that satisfies the assumptions U1-U4, it holds that
\[
\left| \sumi{t}{T} \expec[f_t^\infty(\valpha^*, \vbeta^*)] - \sumi{t}{T} \expec[f_t(\valpha^*, \vbeta^*)] \right| = O(1)
\]
\end{lemma}

\begin{proof}
Let $(\valpha', \vbeta')$ denote the parameters that generated the signal. Thus,
\[
f_t(\valpha', \vbeta') = \ell_t(x_t, x_t - \veps_t)
\]
for all $t$. Since $\veps_t$ is independent of $\veps_1, \ldots, \veps_{t-1}$, the best prediction at time $t$ will cause a loss of at least $\expec[\ell_t(x_t, x_t - \veps_t)].$ Since $\expec[\veps_t] = 0$ and $\ell_t$ is convex, it follows that $(\valpha^*, \vbeta^*) = (\valpha', \vbeta')$ and that
\[
f_t(\valpha^*, \vbeta^*) = \ell_t(x_t, x_t - \veps_t)
\]

We define a few things first. Let 
\[
y_t = \tau\left( x_t \right) - \tau\left( x_t^\infty(\valpha^*, \vbeta^*) \right) - \veps_t,\ \ \ \ \vy_t = \colvec{y_t \\ y_{t-1} \\ \vdots \\ y_{t-q+1}}
\]
WLOG (and by assumption), we can assume that $\expec\left[\|\vy_0\|_2\right] \leq \rho$, where $\rho$ is some positive constant.
Next we show that
\[
\expec[|y_t|] = \expec\left[|\tau\left( x_t \right) - \tau\left( x_t^\infty(\valpha^*, \vbeta^*) \right) - \veps_t| \right] \leq \kappa \lambda_{\max}^{t} \rho
\]
We have that
\bal
\tau\left( x_t \right) - \tau\left( x_t^\infty(\valpha^*, \vbeta^*) \right) - \veps_t =&  \sumi{i}{l_a} \alpha_i^* \tau\left( x_{t-i} \right) + \sumi{i}{l_m} \beta_i^* \veps_{t-i} + \veps_t\\
&- \sumi{i}{l_a}\alpha_i^* \tau\left( x_{t-i} \right) - \sumi{i}{l_m} \beta_i^*\left( \tau\left( x_{t-i} \right) - \tau\left( x_{t-i}^\infty(\valpha^*, \vbeta^*) \right) \right) - \veps_t \\
=&  -\sumi{i}{l_m} \beta_i^*\left( \tau\left( x_{t-i} \right) - \tau\left( x_{t-i}^{\infty}(\valpha^*, \vbeta^*) \right) - \veps_{t-i} \right)
\eal
which shows that $y_t = -\sumi{i}{l_m}\beta_i^* y_{t-i}$. The companion matrix to this difference equation is exactly $\vF$ as defined in Eq. \ref{eq:companion_matrix}. Thus,
\[
\vy_t = \vF \vy_{t-1} 
\]
Next, we note that
\bal
|y_t| &\leq \|\vy_t\|_2 = \|
\vF \vy_{t-1}\|_2\\
&= \|\vF^2 \vy_{t-2} \|_2\\
&= \|\vF^t \vy_0 \|_2\\
&= \|\vT \Lambda^t \vT^{-1} \vy_0\|_2\\
&\leq \|\vT\|_2 \|\vT^{-1}\|_2 \|\Lambda^t\|_2 \|\vy_0 \|_2\\
&= \frac{\sigma_{\max}(\vT)}{\sigma_{\min}(\vT)} \lambda_{\max}^t \|\vy_0\|_2\\
&\leq \kappa \lambda_{\max}^t \|\vy_0\|_2
\eal
Taking the expectation gives us $\expec[|y_t|] \leq \kappa \lambda_{\max}^{t} \expec\left[\|\vy_{0}\|_2\right] \leq \kappa \lambda_{\max}^{t}\rho.$
\\
\\
Now we combine this with the Lipschitz continuity of $\ell_t$ to get
\bal
\left| \expec \left[ f_t^\infty(\valpha^*, \vbeta^*) \right] - \expec \left[ f_t(\valpha^*, \vbeta^*) \right] \right| &= \left| \expec \left[ \ell_t(x_t, x_t^\infty(\valpha^*, \vbeta^*)) \right] - \expec \left[ \ell_t(x_t, x_t - \veps_t) \right] \right|\\
&\leq \expec\left[|\ell_t(x_t, x_t^\infty(\valpha^*, \vbeta^*)) - \ell_t(x_t, x_t - \veps_t)| \right]\\
&\leq L\cdot \expec\left[ |x_t - x_t^\infty(\valpha^*, \vbeta^*) - \veps_t| \right]\\
&= L\cdot \expec\left[ |\tau\left( x_t \right) - \tau\left( x_t^\infty(\valpha^*, \vbeta^*)\right) - \veps_t| \right]\\
&\leq \kappa L\rho \lambda_{\max}^{t}
\eal
where we used Jensen's inequality in the first inequality. Note that we also assume $x_t - \tilde{x}_t = \zeta(\tau(x_t)) - \zeta(\tau(\tilde{x}_t)) = \tau(x_t) - \tau(\tilde{x}_t)$. This holds true for the transformations given in Table \ref{table:dgp}. Summing this from $t=1$ to $T$ gives us the result.
\end{proof}

\begin{lemma} \label{lemma:ftm_to_ftinf}
For any data sequence $\{x_t\}$ that satisfies the assumptions U1-U4, it holds that
\[
\left| \sumi{t}{T}\expec\left[ f_t^m(\valpha^*, \vbeta^*) \right] - \sumi{t}{T}\expec\left[ f_t^\infty(\valpha^*, \vbeta^*) \right] \right| = O(1)
\]
if we choose $m = \log_{\lambda_{\max}}\left((2\kappa TLM_{\max} \sqrt{l_m})^{-1}\right).$
\end{lemma}

\begin{proof}
Fix $t$. Note that for $m < 0$,
\bal
|\tau\left( x_t^m(\valpha^*, \vbeta^*) \right) - \tau\left( x_t^\infty(\valpha^*, \vbeta^*) \right)| &= |\tau\left( x_t \right) - \tau\left( x_t^\infty(\valpha^*, \vbeta^*) \right)|\\
&\leq |\tau\left( x_t \right) - \tau\left( x_t^\infty(\valpha^*, \vbeta^*) \right) - \veps_t| + |\veps_t|
\eal
The right hand side of the inequality is simply $|y_t| + |\veps_t|$, where $y_t$ is as defined in Lemma \ref{lemma:ftinf_to_ft}. By assumption, $\expec[|\veps_t|] < M_{\max}$. Assume that $M_{\max}$ is large enough such that $\expec[|y_t|] \leq M_{\max}$. This is a valid assumption since it is decaying exponentially as proved in Lemma \ref{lemma:ftinf_to_ft}. It is important to note that $\tau\left( x_t^m(\valpha, \vbeta) \right)$ and $\tau\left( x_t^\infty(\valpha, \vbeta) \right)$ have no randomness in them since $\tau$ is deterministic. Thus, for $m < 0$, 
\bal
|\tau\left( x_t^m(\valpha^*, \vbeta^*) \right) - \tau\left( x_t^\infty(\valpha^*, \vbeta^*) \right)| &= \expec\left[|\tau\left( x_t^m(\valpha^*, \vbeta^*) \right) - \tau\left( x_t^\infty(\valpha^*, \vbeta^*)\right)| \right] \\
&= \expec\left[|\tau\left( x_t \right) - \tau\left( x_t^\infty(\valpha^*, \vbeta^*) \right)| \right]\\
&\leq \expec[|y_t| + |\veps_t|] \\
&\leq 2 M_{\max}
\eal
Squaring both sides of the inequality results in 
\[
(\tau\left( x_t^m(\valpha^*, \vbeta^*) \right) - \tau\left( x_t^\infty(\valpha^*, \vbeta^*) \right))^2 \leq 4 M_{\max}^2
\]
Next, we define 
\[
z_t^m = \tau\left( x_t^m(\valpha^*, \vbeta^*) \right) - \tau\left( x_t^\infty(\valpha^*, \vbeta^*) \right),\ \ \ \ \vz_t^m = \colvec{z_t^m \\ z_{t-1}^{m-1} \\ \vdots \\ z_{t-q+1}^{m-q+1}}
\]
We have that
\bal
\tau\left( x_t^m(\valpha^*, \vbeta^*) \right) - \tau\left( x_t^\infty(\valpha^*, \vbeta^*) \right) =& \sumi{i}{l_a} \alpha^*_i \tau\left( x_{t-i} \right) + \sumi{i}{l_m}\beta_i^*(\tau\left( x_{t-i} \right) - \tau\left( x_{t-i}^{m-i}(\valpha^*, \vbeta^*) \right))\\
&- \sumi{i}{l_a}\alpha^*_i \tau\left( x_{t-i} \right) - \sumi{i}{l_m}\beta^*_i(\tau\left( x_{t-i} \right) - \tau\left(x_{t-i}^{\infty}(\valpha^*, \vbeta^*) \right))\\
=& -\sumi{i}{l_m} \beta^*_i \left( \tau\left( x_{t-i}^{m-i}(\valpha^*, \vbeta^*) \right) - \tau\left(x_{t-i}^\infty(\valpha^*, \vbeta^*)\right) \right)
\eal
Thus, $z_t^m = -\sumi{i}{l_m} \beta^*_i z_{t-i}^{m-i}$. The companion matrix to this difference equation is exactly $\vF$ as defined in Eq. \ref{eq:companion_matrix}. Thus,
\[
\vz_t^m = \vF \vz_{t-1}^{m-1}
\]
We have that
\bal
|z_t^m| &\leq \|\vz_t^m\|_2 = \|\vF\vz_{t-1}^{m-1}\|_2\\
&= \| \vF^2 \vz_{t-2}^{m-2}\|_2\\
&= \| \vF^m \vz_{t-m}^{0}\|_2\\
&= \| \vT \Lambda^m \vT^{-1} \vz_{t-m}^0 \|_2\\
&\leq \|\vT \|_2 \|\vT^{-1}\|_2 \|\Lambda^m\|_2 \|\vz_{t-m}^0\|_2\\
&= \frac{\sigma_{\max}(\vT)}{\sigma_{\min}(\vT)} \lambda_{\max}^m \sqrt{\sum_{i=0}^{l_m-1} (z_{t-m-i}^{-i})^2}\\
&\leq \kappa \lambda_{\max}^m \sqrt{q 4 M_{\max}^2}\\
&= \kappa \lambda_{\max}^m  2 M_{\max} \sqrt{l_m}
\eal
Now we combine this with the Lipschitz continuity of $\ell_t$ to get
\bal
\left| \expec[f_t^m(\valpha^*, \vbeta^*)] - \expec[f_t^\infty(\valpha^*, \vbeta^*)] \right| &= \left| \expec[\ell_t(x_t, x_t^m(\valpha^*, \vbeta^*))] - \expec[\ell_t(x_t, x_t^\infty(\valpha^*, \vbeta^*))] \right|\\
&\leq \expec[\left | \ell_t(x_t, x_t^m(\valpha^*, \vbeta^*)) - \ell_t(x_t, x_t^\infty(\valpha^*, \vbeta^*)) \right| ]\\
&\leq L \cdot \expec[|x_t^m(\valpha^*, \vbeta^*) - x_t^\infty(\valpha^*, \vbeta^*)|]\\
&= L \cdot |\tau\left( x_t^m(\valpha^*, \vbeta^*)\right) - \tau\left( x_t^\infty(\valpha^*, \vbeta^*) \right)|\\
&\leq 2 \kappa L M_{\max} \sqrt{l_m} \lambda_{\max}^m
\eal
where in the first inequality we used Jensen's inequality and we again used the assumption that $x_t - \tilde{x}_t = \tau(x_t) - \tau(\tilde{x}_t)$.
\\
\\
Summing this quantity from $t=1$ to $T$ gives us the result:
\[
\left| \sumi{t}{T} \expec[f_t^\infty(\valpha^*, \vbeta^*)] - \sumi{t}{T} \expec[f_t^m(\valpha^*, \vbeta^*)] \right| \leq 2 \kappa TLM_{\max}\sqrt{l_m} \lambda_{\max}^m
\]
Choosing $m = \log_{\lambda_{\max}}\left((2\kappa TLM_{max} \sqrt{l_m})^{-1}\right)$ gives us the desired $O(1)$ property.
\end{proof}

\section{Proof of Theorem \ref{thm:ec_varma_ogd}}

\begin{proof}
We again produce a proof of very similar structure to \citet{AnavaHazan} and \citet{liu2016online}. We first need to redefine a few things for the vector case. Let $ D = \sup_{\vgamma_1, \vgamma_2 \in \mathcal{K}} \|\vgamma_1 - \vgamma_2\|_F$, and $\|\nabla_{\vgamma} \ell_t^m(\vgamma)\|_F \leq G(T)$. 

\textbf{Step 1}: Since $\{\ell_t^M\}$ are convex loss functions, we may invoke \citep{Zinkevich2003OnlineCP} with a fixed step size $\eta = \frac{D}{G(T)\sqrt{T}}$:
\[
\sumi{t}{T} \ell_t^M(\vgamma_t) - \min_{\vgamma} \sumi{t}{T} \ell_t^M(\vgamma) = O\left( DG(T) \sqrt{T} \right)
\]
Again, we note that the proof in \citep{Zinkevich2003OnlineCP} uses a constant upper bound $G$ on the gradients. Since we assume $G(T)$ is a monotonically increasing function, the proof in \citep{Zinkevich2003OnlineCP} follows through straightforwardly.

\textbf{Step 2}: Next we define a few things in the same vein as in the proof of Theorem \ref{thm:tsp_ogd}. Let
\bal
\Delta \vx_t^\infty(\Pi, \Gamma, \vTheta) &= \Pi \vx_{t-1} + \sumi{i}{p-1} \Gamma_i \Delta \vx_{t-i} + \sumi{i}{q} \vTheta_i \left( \Delta \vx_{t-i} - \Delta \vx_{t-i}^\infty (\Pi, \Gamma, \vTheta) \right)\\
\vx_t^\infty(\Pi, \Gamma, \vTheta) &= \Delta \vx_t^\infty (\Pi, \Gamma, \vTheta) + \vx_{t-1}\\
f_t^\infty (\Pi, \Gamma, \vTheta) &= \ell_t\left(\vx_t,\ \vx_t^\infty \left(\Pi, \Gamma, \vTheta\right)\right)
\eal
with initial condition $\Delta \vx_t^\infty (\Pi, \Gamma, \vTheta) = \Delta \vx_t$ for all $t < 0$. Note that we are assuming that we have fixed data $\vx_{0}, \ldots, \vx_{-p}$. With this definition, we can write $\Delta \vx_t^\infty(\Pi, \Gamma, \vTheta) = c_0(\Pi, \Gamma, \vTheta)\vx_{t-1} +  \sumi{i}{t+p-1} c_i(\Pi, \Gamma, \vTheta) \Delta \vx_{t-i}$, i.e. as a growing AR process. This is because we can undo the reparameterization and write $\Delta \vx_t$ in its original VARMA process form
\bal
\vx_t^\infty(\Pi, \Gamma, \Theta) &= \sumi{i}{p} \vA_i \vx_{t-i} + \sumi{i}{q} \vTheta_i \left(\vx_{t-i} - \vx_{t-i}^\infty(\Pi, \Gamma, \Theta)\right)\\
&= \sumi{i}{t+p}c_i(\vA, \Theta) \vx_{t-i}
\eal
as shown in the proof of Algorithm \ref{alg:tsp_ogd}. Using the error corrected reparameterization here results in 
\[
\Delta \vx_t^\infty(\Pi, \Gamma, \Theta) = c_0(\Pi, \Gamma, \Theta) \vx_{t-1}+ \sumi{i}{t+p-1} c_i(\Pi, \Gamma, \Theta) \Delta \vx_{t-i}
\]
Furthermore, we define
\bal
\Delta \vx_t^m(\Pi, \Gamma, \vTheta) &= \Pi \vx_{t-1} + \sumi{i}{p-1} \Gamma_i \Delta \vx_{t-i} + \sumi{i}{q} \vTheta_i \left( \Delta \vx_{t-i} - \Delta \vx_{t-i}^{m-i} (\Pi, \Gamma, \vTheta) \right)\\
\vx_t^m(\Pi, \Gamma, \vTheta) &= \Delta \vx_t^m (\Pi, \Gamma, \vTheta) + \vx_{t-1}\\
f_t^m (\Pi, \Gamma, \vTheta) &= \ell_t(\vx_t,\ \vx_t^m (\Pi, \Gamma, \vTheta))
\eal
with initial condition $\Delta \vx_t^m(\Pi, \Gamma, \vTheta) = \Delta \vx_t$ for all $m < 0$. We relate $M = m+p-1$. With this definition, we can write $\Delta \vx_t^m(\Pi, \Gamma, \vTheta) = \tilde{c}_0(\Pi, \Gamma, \vTheta) \vx_{t-1} + \sumi{i}{M} \tilde{c}_i(\Pi, \Gamma, \vTheta) \Delta \vx_{t-i}$ by using similar rearrangement arguments as shown above.

Lastly, we define
\[
(\Pi^*, \Gamma^*, \vTheta^*) = \argmin_{\Pi, \Gamma, \vTheta} \sumi{t}{T} \expec [f_t(\Pi, \Gamma, \vTheta)]
\]
Recall that $\vx_t$ is fixed in the expectation.

Lemma \ref{lemma:vec_ltm_to_ftm} gives us that
\[
\min_{\vgamma} \sumi{t}{T} \ell_t^M(\vgamma) \leq \sumi{t}{T} f_t^m(\Pi^*, \Gamma^*, \vTheta^*)
\]
Lemma \ref{lemma:vec_ftm_to_ftinf} says that choosing $m = \log_{\lambda_{\max}}\left(\left( 2\kappa TLM_{\max}\sqrt{q}\right)^{-1}\right)$ results in
\[
\left | \sumi{t}{T} \expec[f_t^m(\Pi^*, \Gamma^*, \vTheta^*)] - \sumi{t}{T}\expec[f_t^\infty(\Pi^*, \Gamma^*, \vTheta^*)] \right| = O(1)
\]
Lemma \ref{lemma:vec_ftinf_to_ft} gives us that
\[
\left | \sumi{t}{T} \expec[f_t^\infty(\Pi^*, \Gamma^*, \vTheta^*)] - \sumi{t}{T}\expec[f_t(\Pi^*, \Gamma^*, \vTheta^*)] \right| = O(1)
\]
Chaining all of these gives us the final result:
\[
\sumi{t}{T} \ell_t^M(\vgamma_t) - \min_{\Pi, \Gamma, \vTheta} \sumi{t}{T} \expec[f_t(\Pi, \Gamma, \vTheta)] = O\left(DG(T) \sqrt{T}\right)
\]

\end{proof}

\begin{lemma} \label{lemma:vec_ltm_to_ftm}
For all $m$ and $\{\vx_t\}$ that satisfies assumptions M1-M3, we have that
\[
\min_{\vgamma} \sumi{t}{T} \ell_t^m(\vgamma) \leq \sumi{t}{T} f_t^m(\Pi^*, \Gamma^*, \vTheta^*)
\]
\end{lemma}
\begin{proof}
Recall that $\vgamma = \{\tilde{\Pi}, \tilde{\Gamma}_i,\ i = 1, \ldots, M\}$ We simply set $\tilde{\Pi}^* = \tilde{c}_0(\Pi^*, \Gamma^*, \vTheta^*),\ \tilde{\Gamma}_i^* = \tilde{c}_i(\Pi^*, \Gamma^*, \vTheta^*)$ and let that be denoted by $\vgamma^*$. Thus, we get $\sumi{t}{T} \ell_t^M(\vgamma^*) = \sumi{t}{T} f_t^m(\Pi^*, \Gamma^*, \vTheta^*)$. Thus, the minimum holds trivially. Note that we assume $\vgamma^* \in \mathcal{E}$.
\end{proof}

\begin{lemma} \label{lemma:vec_ftinf_to_ft}
For any data sequence $\{\vx_t\}_{t=1}^T$ that satisfies assumptions M1-M5, it holds that
\[
\left | \sumi{t}{T} \expec[f_t^\infty(\Pi^*, \Gamma^*, \vTheta^*)] - \sumi{t}{T}\expec[f_t(\Pi^*, \Gamma^*, \vTheta^*)] \right| = O(1)
\]
\end{lemma}

\begin{proof}
We start the proof in the same exact way that Anava does. Let $(\Pi', \Gamma', \vTheta')$ denote the parameters that generated the signal. Thus,
\[
f_t(\Pi', \Gamma', \vTheta') = \ell_t(\vx_t, \vx_t - \bveps_t)
\]
for all $t$. Since $\bveps_t$ is independent of $\bveps_1, \ldots, \bveps_{t-1}$, the best prediction at time $t$ will cause a loss of at least $\expec[\ell_t(\vx_t, \vx_t - \bveps_t)].$ Since $\expec[\bveps_t] = 0$ and $\ell_t$ is convex, it follows that $(\Pi^*, \Gamma^*, \vTheta^*) = (\Pi', \Gamma', \vTheta')$ and that
\[
f_t(\Pi^*, \Gamma^*, \vTheta^*) = \ell_t(\vx_t, \vx_t - \bveps_t)
\]

We define a few things first. Let 
\[
\vy_t = \Delta \vx_t - \Delta \vx_t^\infty(\Pi^*, \Gamma^*, \vTheta^*) - \bveps_t,\ \ \ \ \vY_t = \colvec{\vy_t \\ \vy_{t-1} \\ \vdots \\ \vy_{t-q+1}}
\]
(note the overloading from previous sections) By assumption, we can assume that $\expec\left[\|\vY_0\|_2\right] \leq \rho$, where $\rho$ is some positive constant.
Next we show that
\[
\expec[\|\vy_t\|_2] = \expec\left[\|\Delta \vx_t - \Delta \vx_t^\infty(\Pi^*, \Gamma^*, \vTheta^*) - \bveps_t\|_2 \right] \leq \kappa \lambda_{\max}^{t} \rho
\]
We have that
\bal
\Delta \vx_t - \Delta \vx_t^\infty(\Pi^*, \Gamma^*, \vTheta^*) - \bveps_t =&\ \ \ \Pi^* \vx_{t-1} +  \sumi{i}{p-1} \Gamma_i^* \Delta \vx_{t-i} + \sumi{i}{q} \vTheta_i^* \bveps_{t-i} + \bveps_t\\
&-\Pi^* \vx_{t-1} - \sumi{i}{p-1}\Gamma_i^* \Delta \vx_{t-i} - \sumi{i}{q} \vTheta_i^*\left( \Delta \vx_{t-i} - \Delta \vx_{t-i}^\infty(\Pi^*, \Gamma^*, \vTheta^*) \right) - \bveps_t \\
=&  -\sumi{i}{q} \vTheta_i^*\left( \Delta \vx_{t-i} - \Delta \vx_{t-i}^{\infty}(\Pi^*, \Gamma^*, \vTheta^*) - \bveps_{t-i} \right)
\eal
which shows that $\vy_t = -\sumi{i}{q}\vTheta_i^* \vy_{t-i}$. The companion matrix to this difference equation is $\vF$. Thus,
\[
\vY_t = \vF \vY_{t-1} 
\]
Next, we note that
\bal
\|\vy_t\|_2 &\leq \|\vY_t\|_2 = \|
\vF \vY_{t-1}\|_2\\
&= \|\vF^2 \vY_{t-2} \|_2\\
&= \|\vF^t \vY_0 \|_2\\
&= \|\vT \Lambda^t \vT^{-1} \vY_0\|_2\\
&\leq \|\vT\|_2 \|\vT^{-1}\|_2 \|\Lambda^t\|_2 \|\vY_0 \|_2\\
&= \frac{\sigma_{\max}(\vT)}{\sigma_{\min}(\vT)} \lambda_{\max}^t \|\vY_0\|_2\\
&\leq \kappa \lambda_{\max}^t \|\vY_0\|_2
\eal
Taking the expectation gives us $\expec[\|\vy_t\|_2] \leq \kappa (1-\veps)^{t} \expec\left[\|\vY_{0}\|_2\right] \leq \kappa \lambda_{\max}^{t}\rho.$
\\
\\
Now we combine this with the Lipschitz continuity of $\ell_t$ to get
\bal
\left| \expec \left[ f_t^\infty(\Pi^*, \Gamma^*, \vTheta^*) \right] - \expec \left[ f_t(\Pi^*, \Gamma^*, \vTheta^*) \right] \right| &= \left| \expec \left[ \ell_t(\vx_t, \vx_t^\infty(\Pi^*, \Gamma^*, \vTheta^*)) \right] - \expec \left[ \ell_t(\vx_t, \vx_t - \bveps_t) \right] \right|\\
&\leq \expec\left[|\ell_t(\vx_t, \vx_t^\infty(\Pi^*, \Gamma^*, \vTheta^*)) - \ell_t(\vx_t, \vx_t - \bveps_t)| \right]\\
&\leq L\cdot \expec\left[ \|\vx_t - \vx_t^\infty(\Pi^*, \Gamma^*, \vTheta^*) - \veps_t\|_2 \right]\\
&= L\cdot \expec\left[ \|\Delta \vx_t - \Delta \vx_t^\infty(\Pi^*, \Gamma^*, \vTheta^*) - \veps_t\|_2 \right]\\
&\leq \kappa L\rho \lambda_{\max}^{t}
\eal
where we used Jensen's inequality in the first inequality. Summing this from $t=1$ to $T$ gives us the result.
\end{proof}

\begin{lemma} \label{lemma:vec_ftm_to_ftinf}
For any data sequence $\{x_t\}_{t=1}^T$ that satisfies assumptions M1-M3, it holds that
\[
\left| \sumi{t}{T}\expec\left[ f_t^m(\Pi^*, \Gamma^*, \vTheta^*) \right] - \sumi{t}{T}\expec\left[ f_t^\infty(\Pi^*, \Gamma^*, \vTheta^*) \right] \right| = O(1)
\]
if we choose $m = \log_{\lambda_{\max}}\left((2\kappa TLM_{\max} \sqrt{q})^{-1}\right).$
\end{lemma}

\begin{proof}
Fix $t$. Note that for $m < 0$,
\bal
|\Delta \vx_t^m(\Pi^*, \Gamma^*, \vTheta^*) - \Delta \vx_t^\infty(\Pi^*, \Gamma^*, \vTheta^*)| &= |\Delta \vx_t - \Delta \vx_t^\infty(\Pi^*, \Gamma^*, \vTheta^*)|\\
&\leq |\Delta \vx_t - \Delta \vx_t^\infty(\Pi^*, \Gamma^*, \vTheta^*) - \bveps_t| + |\bveps_t|
\eal
The right hand side of the inequality is simply $\|\vy_t\|_2 + \|\bveps_t\|_2$, where $\vy_t$ is as defined in Lemma \ref{lemma:vec_ftinf_to_ft}. By assumption, $\expec[\|\bveps_t\|_2] < M_{\max}$. Assume that $M_{\max}$ is large enough such that $\expec[\|\vy_t\|_2] \leq M_{\max}$. This is a valid assumption since it is decaying exponentially as proved in Lemma \ref{lemma:vec_ftinf_to_ft}. It is important to note that $\Delta \vx_t^m(\Pi, \Gamma, \vTheta)$ and $\Delta \vx_t^\infty(\Pi, \Gamma, \vTheta)$ have no randomness in them (recall that they can be written as a linear combination of past values of the realized data sequence $\Delta \vx_t$). Thus, for $m < 0$, 
\bal
\|\Delta \vx_t^m(\Pi^*, \Gamma^*, \vTheta^*) - \Delta \vx_t^\infty(\Pi^*, \Gamma^*, \vTheta^*)\|_2 &= \expec\left[\|\Delta \vx_t^m(\Pi^*, \Gamma^*, \vTheta^*) - \Delta \vx_t^\infty(\Pi^*, \Gamma^*, \vTheta^*)\|_2 \right] \\
&= \expec\left[\|\Delta \vx_t - \Delta \vx_t^\infty(\Pi^*, \Gamma^*, \vTheta^*)\|_2 \right]\\
&\leq \expec[\|\vy_t\|_2 + \|\bveps_t\|_2] \\
&\leq 2 M_{\max}
\eal
Squaring both sides of the inequality results in 
\[
\|\Delta \vx_t^m(\Pi^*, \Gamma^*, \vTheta^*) - \Delta \vx_t^\infty(\Pi^*, \Gamma^*, \vTheta^*)\|_2^2 \leq 4 M_{\max}^2
\]
Next, we define 
\[
\vz_t^m = \Delta \vx_t^m(\Pi^*, \Gamma^*, \vTheta^*) - \Delta \vx_t^\infty(\Pi^*, \Gamma^*, \vTheta^*),\ \ \ \ \vZ_t^m = \colvec{\vz_t^m \\ \vz_{t-1}^{m-1} \\ \vdots \\ \vz_{t-q+1}^{m-q+1}}
\]
We have that
\bal
\Delta \vx_t^m(\Pi^*, \Gamma^*, \vTheta^*) - \Delta \vx_t^\infty(\Pi^*, \Gamma^*, \vTheta^*) =&\ \ \ \Pi^* \vx_{t-1} + \sumi{i}{k} \Gamma^*_i \Delta \vx_{t-i} + \sumi{i}{q}\vTheta_i^*(\Delta \vx_{t-i} - \Delta \vx_{t-i}^{m-i}(\Pi^*, \Gamma^*, \vTheta^*))\\
&- \Pi^* \vx_{t-1} - \sumi{i}{k}\Gamma^*_i \Delta \vx_{t-i} - \sumi{i}{q}\vTheta^*_i(\Delta \vx_{t-i} - \Delta \vx_{t-i}^{\infty}(\Pi^*, \Gamma^*, \vTheta^*))\\
=& -\sumi{i}{q} \vTheta^*_i \left( \Delta \vx_{t-i}^{m-i}(\Pi^*, \Gamma^*, \vTheta^*) - \Delta \vx_{t-i}^\infty(\Pi^*, \Gamma^*, \vTheta^*) \right)
\eal
Thus, $\vz_t^m = -\sumi{i}{q} \vTheta^*_i \vz_{t-i}^{m-i}$. The companion matrix to this difference equation is exactly $\vF$ as defined above. Thus,
\[
\vZ_t^m = \vF \vZ_{t-1}^{m-1}
\]
We have that
\bal
\|\vz_t^m\|_2 &\leq \|\vZ_t^m\|_2 = \|\vF\vZ_{t-1}^{m-1}\|_2\\
&= \| \vF^2 \vZ_{t-2}^{m-2}\|_2\\
&= \| \vF^m \vZ_{t-m}^{0}\|_2\\
&= \| \vT \Lambda^m \vT^{-1} \vZ_{t-m}^0 \|_2\\
&\leq \|\vT \|_2 \|\vT^{-1}\|_2 \|\Lambda^m\|_2 \|\vZ_{t-m}^0\|_2\\
&= \frac{\sigma_{\max}(\vT)}{\sigma_{\min}(\vT)} \lambda_{\max}^m \sqrt{\sum_{i=0}^{q-1} \|\vz_{t-m-i}^{-i}\|_2^2}\\
&\leq \kappa \lambda_{\max}^m \sqrt{q 4 M_{\max}^2}\\
&= \kappa \lambda_{\max}^m  2 M_{\max} \sqrt{q}
\eal
Now we combine this with the Lipschitz continuity of $\ell_t$ to get
\bal
\left| \expec[f_t^m(\Pi^*, \Gamma^*, \vTheta^*)] - \expec[f_t^\infty(\Pi^*, \Gamma^*, \vTheta^*)] \right| &= \left| \expec[\ell_t(\vx_t, \vx_t^m(\Pi^*, \Gamma^*, \vTheta^*))] - \expec[\ell_t(\vx_t, \vx_t^\infty(\Pi^*, \Gamma^*, \vTheta^*))] \right|\\
&\leq \expec[\left | \ell_t(\vx_t, \vx_t^m(\Pi^*, \Gamma^*, \vTheta^*)) - \ell_t(\vx_t, \vx_t^\infty(\Pi^*, \Gamma^*, \vTheta^*)) \right| ]\\
&\leq L \cdot \expec[\|\vx_t^m(\Pi^*, \Gamma^*, \vTheta^*) - \vx_t^\infty(\Pi^*, \Gamma^*, \vTheta^*)\|_2]\\
&= L \cdot \|\Delta \vx_t^m(\Pi^*, \Gamma^*, \vTheta^*) - \Delta \vx_t^\infty(\Pi^*, \Gamma^*, \vTheta^*)\|_2\\
&\leq 2 \kappa L M_{\max} \sqrt{q} \lambda_{\max}^m
\eal
where in the first inequality we used Jensen's inequality.
\\
\\
Summing this quantity from $t=1$ to $T$ gives us the result:
\[
\left| \sumi{t}{T} \expec[f_t^m(\Pi^*, \Gamma^*, \vTheta^*)] - \sumi{t}{T} \expec[f_t^\infty(\Pi^*, \Gamma^*, \vTheta^*)] \right| \leq 2 \kappa TLM_{\max}\sqrt{q} \lambda_{\max}^m
\]
Choosing $m = \log_{\lambda_{\max}}\left((2\kappa TLM_{max} \sqrt{q})^{-1}\right)$ gives us the desired $O(1)$ property.
\end{proof}

\section{Proof of Theorem \ref{thm:ftl_data_dependent}}

\begin{proof}
Recall that for FTL, we have that
\[
\vgamma_t \in \argmin_{{\vgamma}} \sumi{i}{t-1} \ell_t(\vgamma) = \argmin_{{\vgamma}} \frac12 \sumi{i}{t-1} (x_t - {\vgamma}^\intercal \vpsi_t )^2 = \argmin_{\vgamma} \frac12 \| X_t - \Psi_t {\vgamma} \|_2^2
\]
where $X_t = \colvec{x_t & \ldots & x_1}^\T, \Psi_t = \colvec{\vpsi_t & \ldots \vpsi_1}^\T$. Note that this is simply a recursive least squares procedure. This procedure can be computed in a recursive manner using the update equations:
\bal
{\vgamma}_{t+1} &= \vgamma_t + \frac{x_t - \vpsi_t^\T \vgamma_t}{1 + \vpsi_t^\T V_{t-1} \vpsi_t} V_{t-1} \vpsi_{t}\\
V_{t+1} &= V_t - \frac{V_t \vpsi_{t+1} \vpsi_{t+1}^\T V_t}{1 + \vpsi_{t+1}^\T V_t \vpsi_{t+1}}
\eal
where $V_t = \left( \sumi{i}{t} \vpsi_i \vpsi_i^\T \right)^{-1}$. Using the fact that $\ell_t$ is Lipschitz, we have that

\bal
|\ell_t(\vgamma_t) - \ell_t({\vgamma}_{t+1})| &\leq L\| {\vgamma}_{t+1} - \vgamma_t \|_2\\
&= L \left \| \frac{x_t - \vgamma_t^\T \vpsi_t}{1 + \vpsi_t^\T V_{t-1} \vpsi_t} V_{t-1} \vpsi_t \right\|_2\\
&\leq L \left|\frac{x_t - \vgamma_t^\T \vpsi_t}{1 + \vpsi_t^\T V_{t-1} \vpsi_t} \right| \|V_{t-1}\|_2 \|\vpsi_t\|_2\\
&\leq L^2  \left  \|V_{t-1} \right\|_2\\
&= L^2 \lambda_{\max}\left( V_{t-1} \right)\\
&= \frac{L^2}{(t-1) \lambda_{\min}(t-1)}
\eal
where we used the fact that $ \left\|\nabla_{\vgamma} \ell_t({\vgamma}) \right\|_2 = |x_t - {\vgamma}^\T \vpsi_t| \|\vpsi_t\|_2 \leq L,\ \frac{1}{1 + \vpsi_t^\T V_{t-1} \vpsi_t} \leq 1$.

To complete the proof, we sum this quantity up and invoke Lemma \ref{lemma:FTL_reg}. To avoid the divide-by-zero, simply start the indexing at $t=2$.

\end{proof}

\begin{lemma} \label{lemma:FTL_reg}
Let $\ell_1, \ldots, \ell_T$ be a sequence of loss functions. Let ${\vgamma}_1, \ldots, \vgamma_t$ be produced by FTL. Then 
\[
\sumi{t}{T} \ell_t(\vgamma_t) - \min_{\vgamma} \sumi{t}{T} \ell_t({\vgamma}) \leq \sumi{t}{T} \left[ \ell_t(\vgamma_t) - \ell_t({\vgamma}_{t+1}) \right]
\]
\end{lemma}

This is fairly standard material. For reference to a proof, see \citep{liangcs229t}.

\section{Proof of Theorem \ref{thm:tsp_ogd_unknown_transformation}}

We first give an extension of the (randomized) Weighted Majority Algorithm to handle unbounded loss. We include the proof for completeness.
\begin{lemma} \label{lemma:rwm_unbounded}
Assume we run the weighted majority algorithm (see \cite{shalev2011online}) with the modified update rule $w_{t+1}(h) = w_t(h) (1-\eta)^{\frac{\ell_t(h)}{b_t}}$, where $b_t = \max_{\tau \in \{t-k, \ldots, t\}, h \in \mathcal{M}} \ell_{\tau}(h)$, and $\ell_t(h) \geq 0$ for all $t, h$. Define the expected loss of the algorithm to be $\ell_t(\text{ALG}) := \expec_{w_t} [ \ell_t(h_t)] = \sum_{h} \frac{w_t(h)}{W_t} \ell_t(h)$, where $W_t := \sum_h w_t(h)$. Then the resulting regret bound is
\[
\text{Regret}_T := \sumi{t}{T} \ell_t(\text{ALG}) - \min_{h \in \mathcal{M}} \sumi{t}{T} \ell_t(h)\ \leq\ 2 B_T \sqrt{T \log n}
\]
where $B_T = \max_{t \in \{1, \ldots T\}} b_t = \max_{t \in \{1, \ldots T\}, h \in \mathcal{M}} \ell_t(h)$. 
\end{lemma}
\begin{proof}
Using the update
\[
w_{t+1}(h) = w_t(h) (1-\eta)^{\frac{\ell_t(h)}{b_t}} \implies w_{t+1}(h) = (1-\eta)^{\sumi{\tau}{t} \frac{\ell_{\tau}(h)}{b_{\tau}}}
\]

Note that ${\frac{\ell_t(h)}{b_t}} \in [0,1]$ by definition of $b_t$. Using this, we have
\bal
W_{T+1} &= \sum_{h \in \mathcal{H}} w_{T+1}(h) =  \sum_{h \in \mathcal{H}} w_T(h)(1-\eta)^{\frac{\ell_T(h)}{b_T}} \\
&\leq \sum_{h \in \mathcal{H}} w_T(h) \left(1 - \eta  \left( {\frac{\ell_T(h)}{b_T}}\right)\right) \\
&= \sum_{h \in \mathcal{H}} w_{T}(h) - \eta \sum_{h \in \mathcal{H}} w_{T}(h)\left( {\frac{\ell_T(h)}{b_T}}\right)\\
&= W_T - \frac{\eta}{b_T} \left( \sum_{h \in \mathcal{H}} w_T(h) \ell_T(h) \right)\\
&= W_T - \frac{\eta}{b_T} W_T \ell_T(\text{ALG}) \\
&= W_T \left( 1 - \frac{\eta}{b_T}\ell_T(\text{ALG}) \right)\\
&\leq W_T e^{- \eta \frac{\ell_T(\text{ALG})}{b_T}} \\
&\leq n e^{- \eta \sumi{t}{T} \frac{\ell_{t}(\text{ALG})}{b_t}}
\eal
We also note that
\bal
W_{T+1} \geq \max_{h \in \mathcal{M}} w_{T+1}(h) = \max_{h \in \mathcal{M}}\ (1-\eta)^{\sumi{t}{T} \frac{\ell_{t}(h)}{b_{t}}} = (1-\eta)^{\min \limits_{h \in \mathcal{M}} \sumi{t}{T} \frac{\ell_t(h)}{b_t}}
\eal
And thus,
\[
(1-\eta)^{\frac{1}{B_T} \min_{h \in \mathcal{M}} \sumi{t}{T} \ell_t(h) } \leq (1-\eta)^{\min \limits_{h \in \mathcal{M}} \sumi{t}{T} \frac{\ell_t(h)}{b_t}} \leq n e^{- \eta \sumi{t}{T} \frac{\ell_{t}(\text{ALG})}{b_t}} \leq ne^{-\frac{\eta}{B_T} \sumi{t}{T} \ell_t(\text{ALG})}
\]
The first and third inequalities are due to the fact that $B_T \geq b_t,\ \forall t \in \{1, \ldots, T\}.$ For concise notation, denote $\ell_{1 \ldots T} (h) := \sumi{t}{T} \ell_t(h)$.

Taking logs and using the fact that $x \leq -\log(1-x) \leq x(1+x)$, we have
\bal
\log(1-\eta) \frac{1}{B_T} \left( \min \limits_{h \in \mathcal{M}}(\ell_{1\ldots T}(h))\right) &\leq \log n - \frac{\eta}{B_T}\ell_{1\ldots T}(\text{ALG})\\
\log(1-\eta) \min \limits_{h \in \mathcal{M}}(\ell_{1\ldots T}(h)) &\leq B_T \log n  - \eta \ell_{1\ldots T}(\text{ALG})\\
- \eta(1+\eta)\min \limits_{h \in \mathcal{M}}(\ell_{1\ldots T}(h))  &\leq B_T \log n - \eta \ell_{1\ldots T}(\text{ALG})\\
- (1+\eta)\min \limits_{h \in \mathcal{M}}(\ell_{1\ldots T}(h))  &\leq \frac{B_T}{\eta} \log n -  \ell_{1\ldots T}(\text{ALG})\\
\text{Regret}_T &\leq \frac{B_T}{\eta} \log n + \eta \min \limits_{h \in \mathcal{M}}(\ell_{1\ldots T}(h))
\eal
Since $\ell_{1\ldots T}(h) \leq B_T T$, 
\bal
\text{Regret}_T &\leq \frac{B_T}{\eta} \log n + \eta B_T T
\eal
Choosing $\eta = \sqrt{\frac{\log n}{T}}$, we get
\[
\text{Regret}_T \leq 2 B_T \sqrt{ T \log n} 
\]
\end{proof}

Our result from Theorem \ref{thm:tsp_ogd} gives us
\begin{equation*}
\sumi{t}{T} \ell_{t,h}^M \left( \vgamma_h^{(t)} \right) - \min_{\valpha, \vbeta \in \mathcal{K}} \sumi{t}{T} \expec\left[ f_t(\valpha, \vbeta) \right] = O\left( D_h G_h(T) \sqrt{T}\right)
\end{equation*}
for all $h \in \mathcal{M}$. Adding these together gives
\begin{adjustwidth}{-0.65in}{}
\bal
\sumi{t}{T} \expec \left[ \ell_t(h_t) \right] - \min_h \sumi{t}{T} \ell_t(h) + \sumi{t}{T} \ell_{t,h^*}^M \left( \vgamma_{h^*}^{(t)} \right) - \min_{\valpha, \vbeta \in \mathcal{K}} \sumi{t}{T} \expec\left[ f_t(\valpha, \vbeta) \right] &= O \left( B_T \sqrt{T \log n} \right) + O\left( D_{h^*} G_{h^*}(T) \sqrt{T}\right)
\eal
\end{adjustwidth}
where we define $h^* := \argmin_h \sumi{t}{T} \ell_t(h)$ for brevity of notation. The middle two terms cancel, and since $n$ is typically very small, we can treat it as a small constant and absorb it into the big O notation. Combined with the definitions of $D_*, G_*(T)$, this leaves us with
\begin{equation*}
\sumi{t}{T} \expec \left[ \ell_t(h_t) \right] - \min_{\valpha, \vbeta \in \mathcal{K}} \sumi{t}{T} \expec\left[ f_t(\valpha, \vbeta) \right]  \leq  O\left(  \max_h\left\{B_T, D_* G_*(T)\right\} \sqrt{T} \right)
\end{equation*}

Regarding Remark \ref{remark:least_squares_loss},  we show that $\ell_t(h) = \ell_{t,h}^M \left(\vgamma^{(t)}_h\right) = O \left(G_h(t)\right)$. In the setting of least squares (which is one of the most widely used loss function in time series):
\[
\ell_{t}^M \left(\vgamma\right) = \frac12 \left ( x_t - \vgamma^\T \vpsi_t \right )^2
\]
The norm of the gradient of this loss is
\[
\left | x_t - \vgamma^\T \vpsi_t \right | \| \vpsi_t \|_2 \leq G(t)
\]
where the bound is by definition of $G(t)$. It's easy to see that $(x_t - \vgamma^\T \vpsi_t) \leq O \left( \| \vpsi_t \|_2 \right) $ when $\mathcal{E}$ is a norm ball. Thus, $\ell_t(h) = O\left(G_h(t)\right)$. Because $B_T, G_h(T)$ are nondecreasing in $T$, it follows that $B_T = O(G_*(T))$.

\newpage
\section{Data for Experiments}
In this section, we display the data we used in Section \ref{sec:empirical_results}. 

\begin{figure*}[ht]
\centering
\subfloat[Turkey electricity demand]{\includegraphics[width=0.32\linewidth]{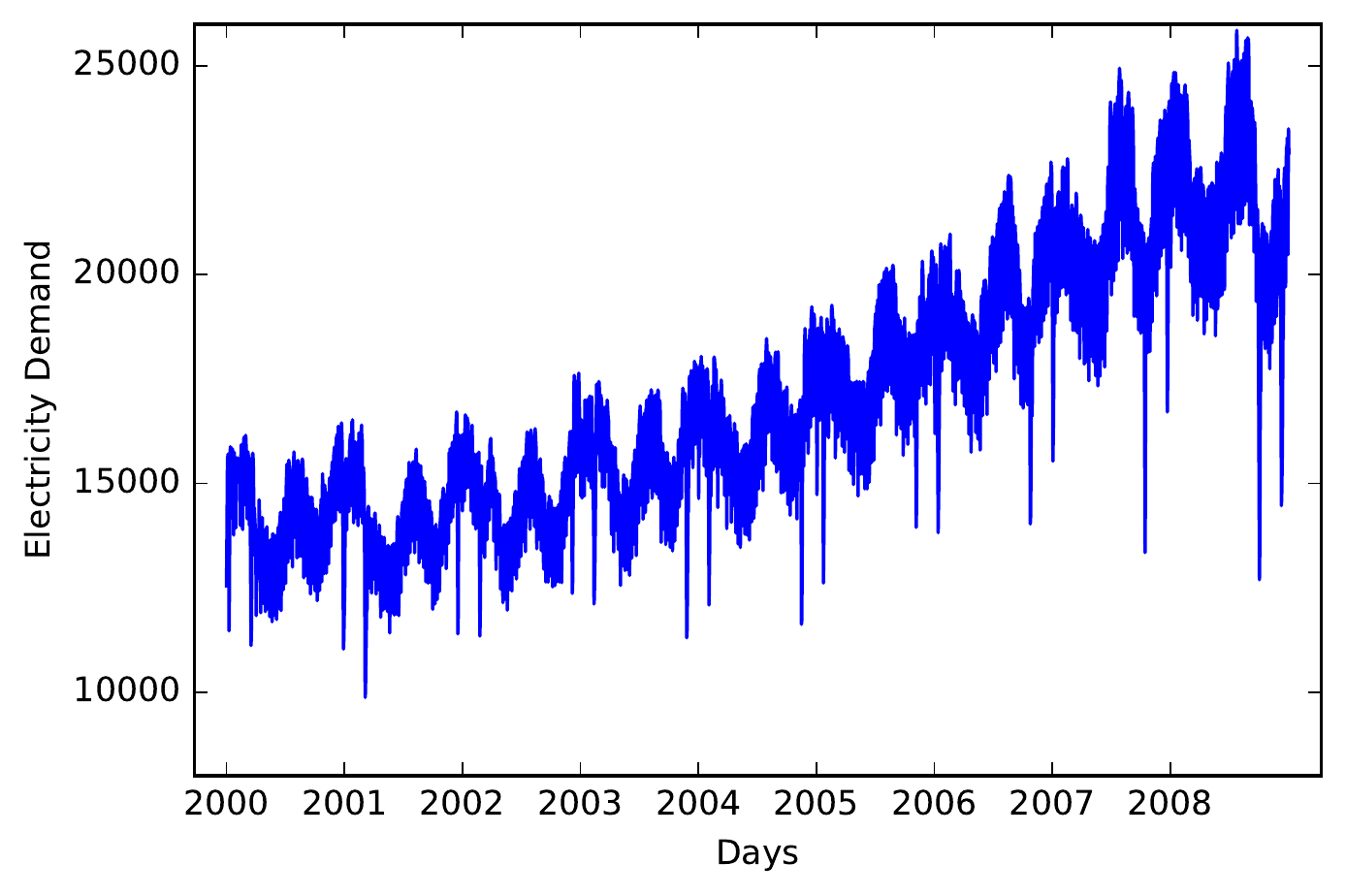}}
\subfloat[Births in Quebec]{\includegraphics[width=0.32\linewidth]{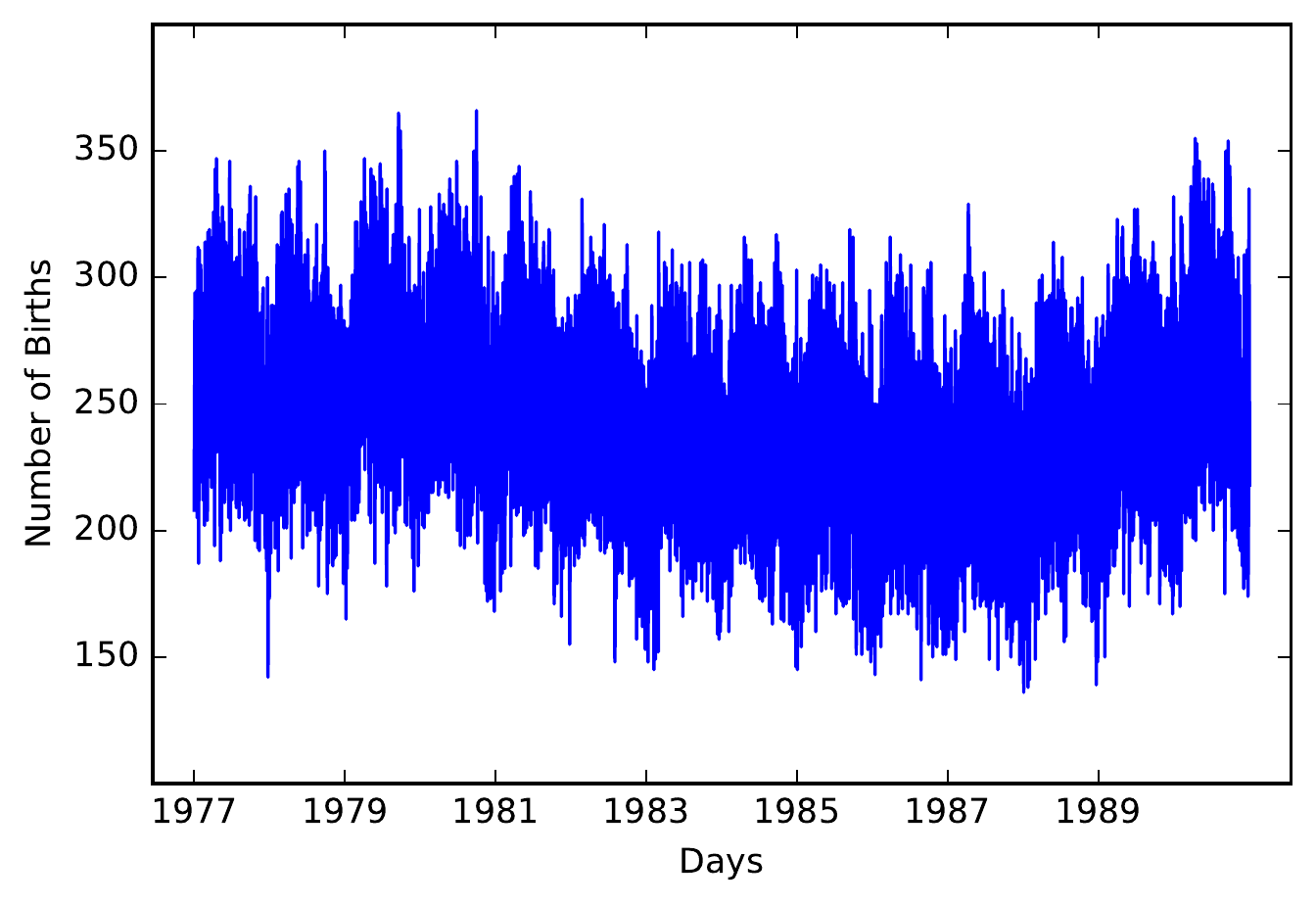}}
\subfloat[Daily River Flows (Saugeen)]{\includegraphics[width=0.32\linewidth]{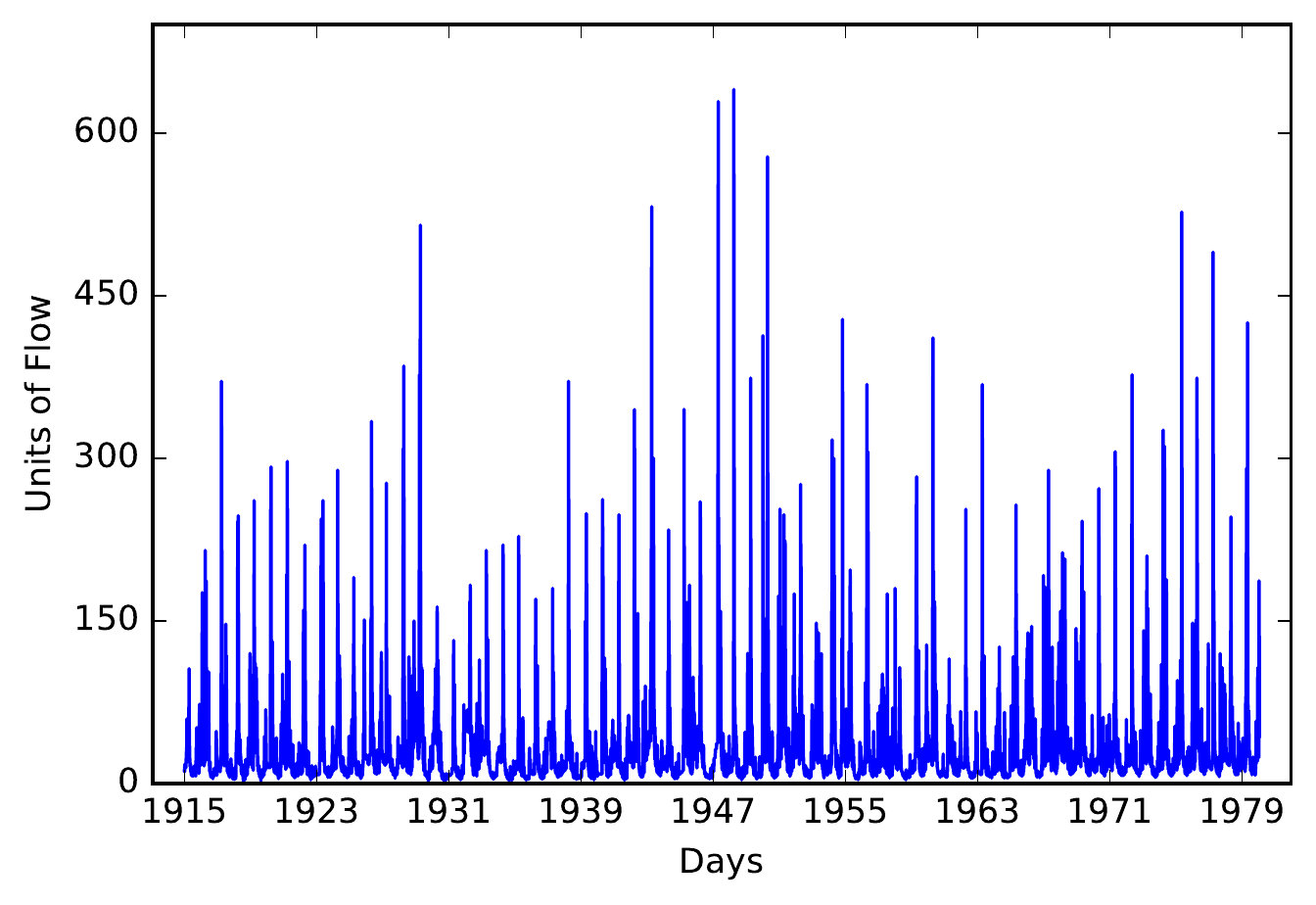}}
\\
\subfloat[Stock data]{\includegraphics[width=0.48\linewidth]{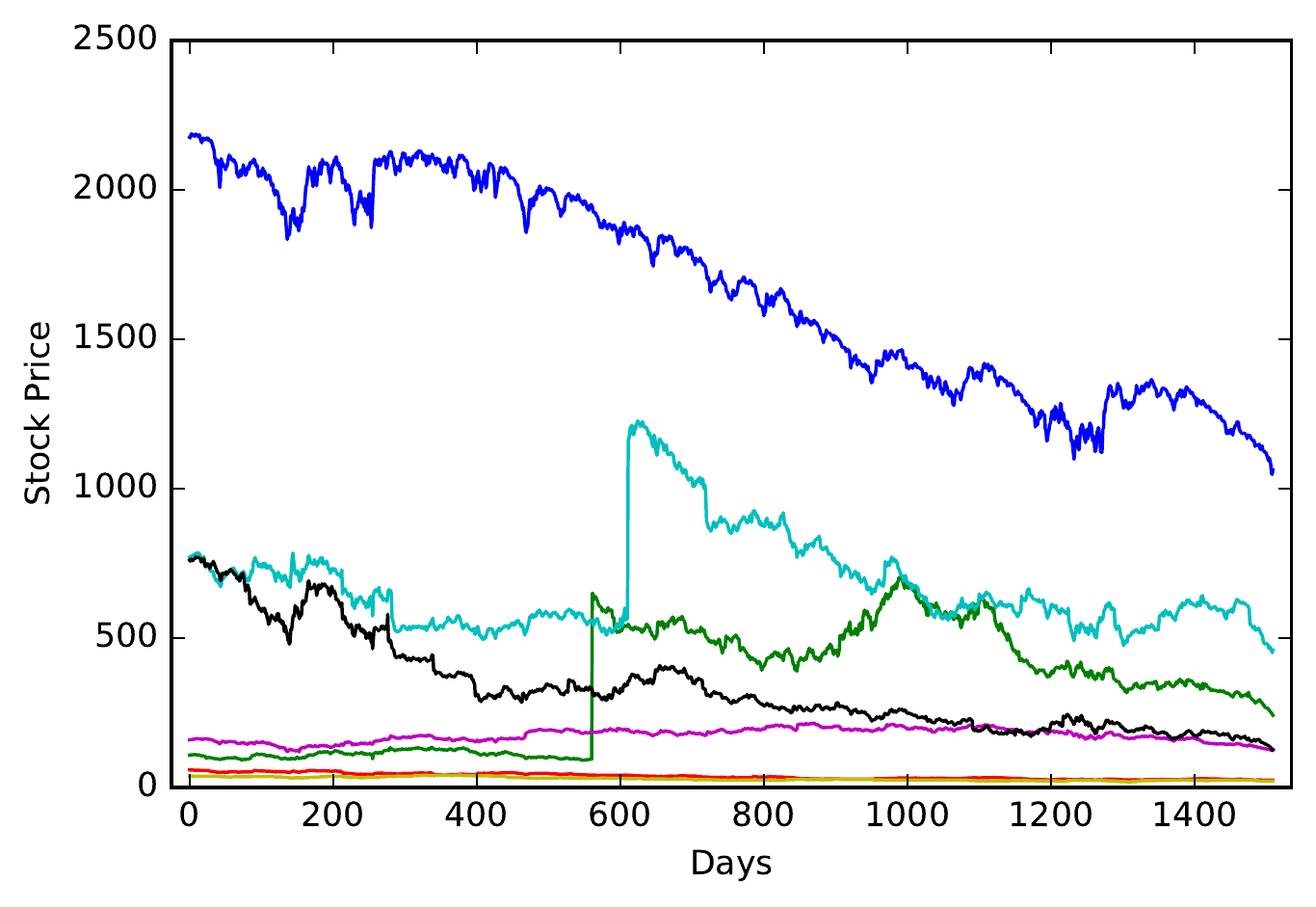}}
\subfloat[Google flu data]{\includegraphics[width=0.48\linewidth]{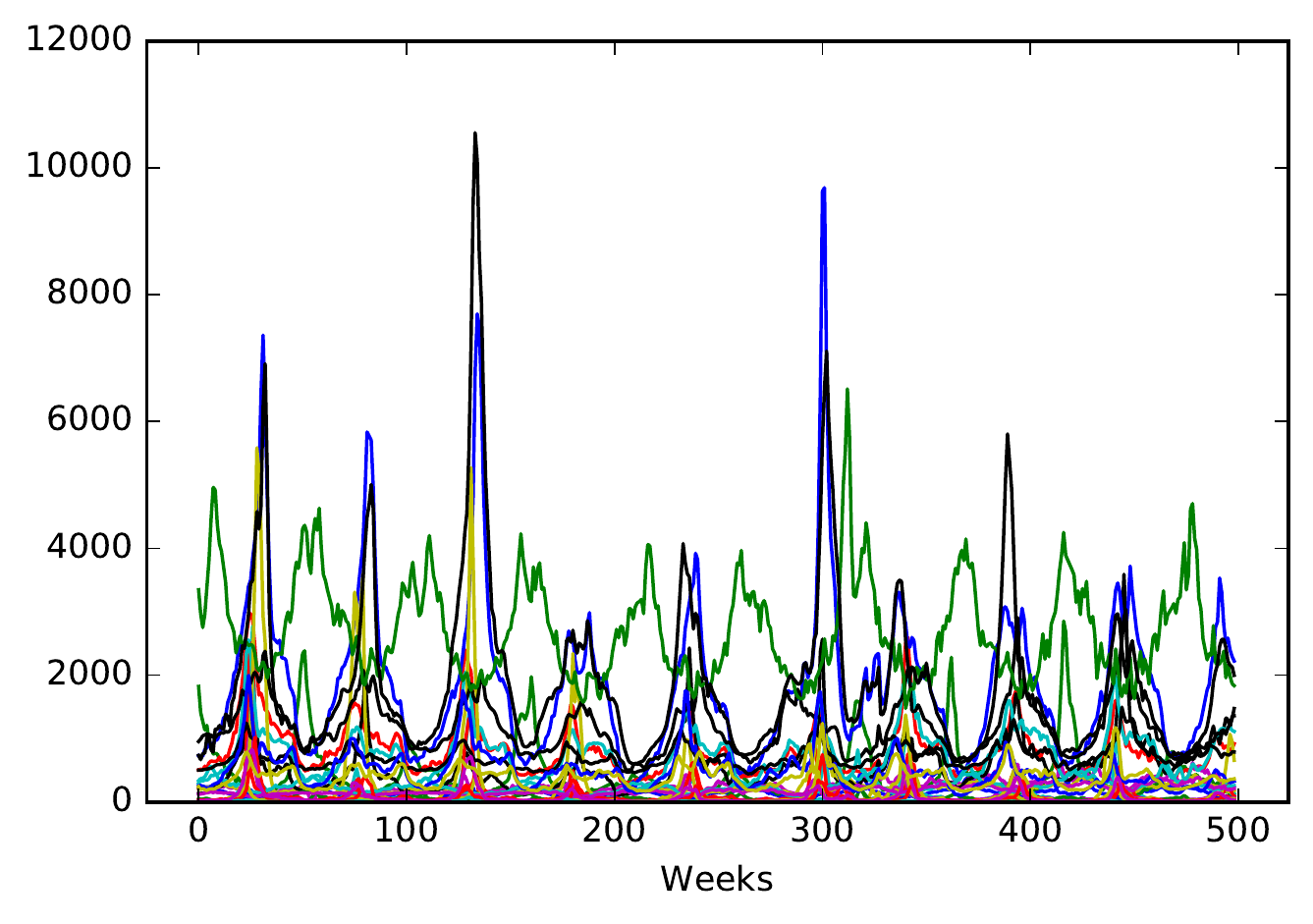}}
\caption{Data plots. The top line has plots for univariate data, and the bottom line has plots for multivariate data.}
\end{figure*}

\end{document}